\documentclass{article}

\usepackage[preprint]{corl_2026} 

\usepackage{pifont}
\usepackage{graphicx}
\usepackage{amsmath,amssymb} 
\usepackage{booktabs}
\usepackage{caption}
\usepackage{makecell}

\usepackage{float}

\usepackage{pifont}
\usepackage{wrapfig}
\usepackage{multirow}
\usepackage{wrapfig}

\usepackage[ruled,vlined]{algorithm2e}
\SetKwInput{KwInput}{Input}
\SetKwInput{KwOutput}{Output}

\usepackage[accsupp]{axessibility}  

\usepackage{caption}
\usepackage{color}
\usepackage[table]{xcolor}
\usepackage{adjustbox}

\definecolor{lightskyblue}{RGB}{230, 230, 255}
\definecolor{lightgrey}{RGB}{237, 237, 237}
\definecolor{lightgreen}{RGB}{90, 200, 100}
\definecolor{lightred}{RGB}{255, 4, 7}
\definecolor{redred}{RGB}{255, 230, 150}
\definecolor{greengreen}{RGB}{150, 255, 150}
\newcommand{\ours}[0]{\rowcolor{lightgrey}}

\newcommand{\ModelName}{FlowPilot}
\newcommand{\ModelNameIL}{FlowPilot-Base}
\newcommand{\ModelNameFull}{FlowPilot-HP}
\usepackage{wrapfig}   
\usepackage{graphicx}  
\usepackage{booktabs}  
\usepackage{caption}   

\newcommand{\Dataset}{SideWalks-300}

\usepackage{hyperref}
\newcommand{\eg}{\textit{e.g.}}
\newcommand{\ie}{\textit{i.e.}}
\usepackage{enumitem}

\title{From Imitation to Alignment: Human-Preference Flow Policies for Long-Horizon Sidewalk Navigation}

\author{
Honglin He \qquad
Zhizheng Liu \qquad
Yukai Ma \qquad
Bolei Zhou\\
University of California, Los Angeles\\
\url{https://vail-ucla.github.io/FlowPilot} \\
}

\begin{document}
\maketitle

\vspace{-5mm}
\begin{abstract}
Autonomous long-horizon sidewalk navigation is essential for micro-mobility applications such as robotic food delivery and assistive electronic wheelchairs. 
Unlike autonomous driving on the road, long-horizon sidewalk navigation requires precise maneuvering through unpredictable sidewalk terrains and pedestrians, with a lightweight perception stack as minimal as a single monocular RGB camera. 
While imitation learning (IL) from demonstrations offers a practical solution, the resulting autopilot policy often suffers from compounding errors, a lack of social compliance on sidewalks, and deficiencies in counterfactual reasoning to handle complex situations.
To address these challenges, we introduce \ModelName, a mapless navigation policy that achieves robust and efficient long-horizon navigation performance using only a monocular RGB camera.
We propose to use anchored flow matching as an action representation for policy pre-training on large-scale robot fleet data and to capture the diverse, complex, multimodal distribution of sidewalk navigation behaviors. To bridge the gap between imitation and alignment, we further design a human-in-the-loop preference learning scheme to tune the policy on a small amount of human intervention data. It strengthens the model's counterfactual reasoning and social compliance on sidewalks.
%
We evaluate \ModelName\ through simulation and real-world experiments in diverse sidewalk environments. \ModelNameIL\ achieves 42\% success rate and 66\% route completion in simulation, while \ModelNameFull\ further improves robustness and social compliance, reducing IR by 40.0\% and NIR by 52.1\% relative to the base model.
%
\end{abstract}


\thispagestyle{plain}
\section{Introduction}
\label{intro}

\begin{figure}[t!]
\centering
\includegraphics[width=\columnwidth]{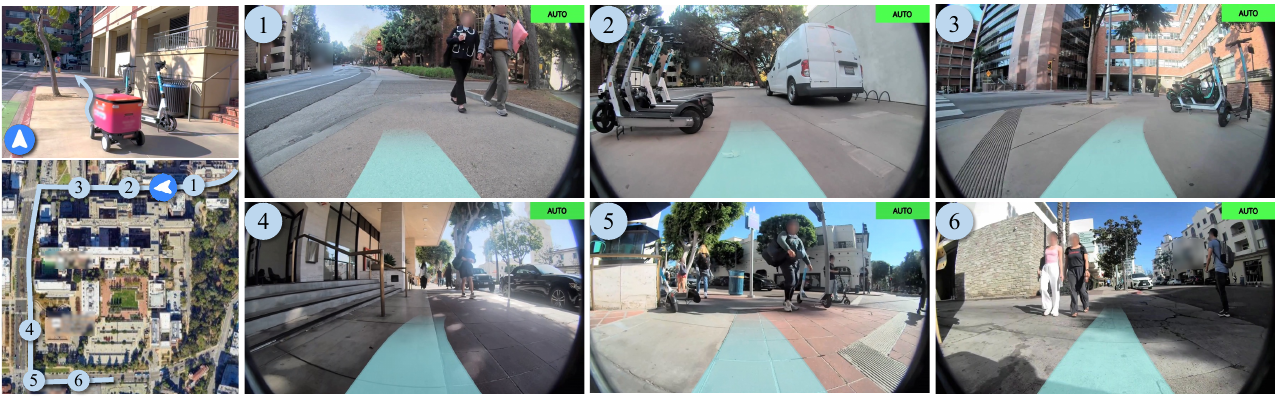} 
\caption{\textbf{Long-horizon sidewalk navigation} for a food delivery robot (the pink bot shown in the upper-left corner). Equipped with only a monocular camera and a GPS, the wheeled robot navigates complex sidewalk terrains, avoid pedestrians and obstacles, and follow social norms to reach its destination safely. We show the first-person views of the robot and the trajectories from \ModelName.
}
\vspace{-8mm}
\label{fig:teaser}
\end{figure}

%
An increasing number of mobile robots are being deployed on urban sidewalks to perform everyday tasks such as food delivery and personal mobility assistance~\cite{engesser2023autonomous, liu2025service, tuomi2021applications}. 
As shown in Fig.~\ref{fig:teaser}, a food delivery robot must navigate a couple of miles along sidewalks to reach its destination. It needs to safely avoid collisions with obstacles like scooters and interact socially with pedestrians. This setting also imposes heavy onboard computing and battery constraints, requiring a lightweight perception stack as minimal as just a single monocular camera. Furthermore, sidewalk layouts can vary significantly across cities and even neighborhoods. Local conditions also change frequently due to construction, temporary obstacles such as abandoned furniture and parked scooters, and unpredictable pedestrian activity, creating significant challenges for autonomous sidewalk navigation~\citep{arntz2023assessment}.

Recent advances in end-to-end navigation policy learning~\cite{shah2022gnm,shah2023vint,hu2023planning,sridhar2024nomad,liu2024citywalker} offer a promising way to directly map visual observations to actions, enabling urban navigation without explicit mapping or heavyweight sensors. Most policies are trained with imitation learning (IL)~\cite{argall2009survey} on large-scale offline data, leveraging demonstrations from heterogeneous sources, including teleoperated robot fleets~\cite{hirose2025learning,karnan2022socially}, human demonstrations~\cite{pan2020zero}, and web videos~\cite{liu2024citywalker}. It has enabled increasingly capable visual navigation policies and has shown potential for cross-embodiment generalization.

Despite these advances, the mapless long-horizon sidewalk navigation remains challenging.
First, standard IL often lacks an expressive action representation. Sidewalk navigation is inherently multi-modal and control-sensitive, \eg, the robot may pass an obstacle on either side, yield or proceed at an intersection, and continuously adjust its turning and speed. However, common action parameterizations are limited. Unimodal Gaussians are too restrictive, while diffusion or flow-matching policies~\cite{chi2025diffusion,sridhar2024nomad} may introduce excessive randomness and temporal inconsistency, and anchor-based regression methods~\cite{he2025seeing} often oversmooth fine-grained variations. To address this, we introduce anchored flow matching, which combines discrete behavior priors with continuous flow refinement to preserve multimodal coverage while generating smooth, temporally consistent future trajectories.

On the other hand, scaling model training on more diverse data does not necessarily yield better, more precise maneuvering behaviors. We observe that imitation learning on web data or robot fleet data creates a \textit{generalization-precision dilemma}: while training on diverse data improves the policy's generalization across environments and robot instances, it may compromise the precision of the policy's action output when deployed on a specific robot instance, where each instance might have subtle variations in camera parameters, mechanics, and physical attributes. Because sidewalk navigation requires high precision, it is essential to further align the navigation model with the specific robot instance on which it is deployed. In the literature, interactive imitation learning~\cite{ross2011reduction,celemin2022interactive,peng2023learning,cai2025predictive} mitigates distribution shift by collecting on-policy corrective supervision, while preference learning shows that comparative human feedback can steer model behavior without dense expert demonstrations~\cite{ziegler2019fine,ouyang2022training,rafailov2023direct}. Inspired by these insights, we design a human-in-the-loop finetuning scheme after large-scale imitation pretraining, bridging large-scale imitation and robot-specific alignment.

In this work, we propose a learning framework for robust long-horizon sidewalk navigation that supports multi-modal action prediction and model alignment from human feedback. We first introduce an anchored flow-matching policy to capture complex multimodal action distributions under long-horizon intentions. To incorporate goal information without inducing shortcut behaviors, we further rank and select candidate actions via a gated conditioning mechanism, encouraging the policy to ground decisions in scene context rather than directly taking the shortest path.
To further adapt a pretrained policy to the precise control requirements of specific robots in complex sidewalk environments, we develop an on-policy preference-learning algorithm using interventions collected during deployment. Specifically, we fine-tune the policy with interventions and preference supervision tailored to our anchored action representation, increasing the likelihood of corrected actions while suppressing undesirable executed actions. This enables a smooth transition from large-scale pretraining to test-time alignment for specific robots. We summarize our contributions:

1) We propose an anchored flow-matching policy to model action distributions for long-horizon sidewalk navigation with gated conditioning to mitigate goal-driven shortcuts.

2) We introduce a reward-free preference learning algorithm that aligns the policy with socially compliant behaviors from human feedback while preserving imitation priors.

3) We evaluate our method through extensive simulation and real-world experiments in diverse and challenging sidewalk environments. Our method achieves 42\% success rate and 66\% route completion in closed-loop simulation benchmark, and improves real-world robustness and social compliance, reducing IR by 40.0\% and NIR by 52.1\% after preference fine-tuning.
\vspace{-2mm}
\section{Related Work}
\vspace{-2mm}

\paragraph{Long-horizon sidewalk navigation}
Visual navigation has been extensively studied, but task difficulty varies significantly with environmental scale and dynamics, map access, robot perception, and computing constraints. Traditional approaches build maps for localization and planning~\cite{thrun2002probabilistic}, but they are difficult to maintain in sidewalks where layouts change, dynamic agents are pervasive, and reliable mapping is costly. Recent work increasingly adopts end-to-end visuomotor policies that map raw sensory observations directly to waypoints~\cite{shah2023gnm,shah2023vint,sridhar2024nomad,liu2025citywalker,chen2025socialnav,he2025seeing,he2026learning}. However, many works focus on indoor or small-scale outdoor navigation~\cite{shah2023gnm,sridhar2024nomad,cai2025navdp}, where horizons are shorter and interactions are constrained. In contrast, sidewalk navigation requires sustained goal pursuit under dense obstacles, pedestrians, and implicit social norms, making compounding errors and social compliance central challenges~\cite{chen2025socialnav}. Although VLMs can enhance reasoning~\cite{chen2025socialnav,cheng2024navila,wei2025ground}, deploying VLMs on compact sidewalk robots remains difficult due to strict latency, compute, and battery constraints.

\vspace{-2mm}
\paragraph{Navigation foundation models}
Recent navigation foundation models aim to improve generalization by scaling training data across environments, and robots~\cite{shah2023vint,bar2024navigation,liu2025citywalker,he2025seeing,bar2024navigation,wei2025ground, chen2025socialnav}. By leveraging large-scale visual data, these models can reduce the need for task-specific training. These advances suggest that large-scale pretraining provides a strong initialization for sidewalk navigation. However, scaling passive demonstrations alone does not fully address the closed-loop nature of long-horizon navigation. Offline data often lacks interactive recovery behaviors~\cite{he2026learning}, and human corrections in safety-critical states. As a result, policies trained primarily through imitation may still suffer from distributional shift, shortcut behaviors, and limited causal understanding~\cite{chen2025socialnav}. 

\vspace{-2mm}
\paragraph{Human alignment for navigation}
Safe and socially compliant sidewalk navigation requires more than imitating expert trajectories; the policy must also align with social norms and deployment-specific safety requirements. Early approaches such as inverse reinforcement learning and adversarial imitation learning~\cite{kretzschmar2016socially,ho2016generative,ziebart2008maximum} infer reward functions~\cite{seneviratne2025halo} from demonstrations, but they are computationally expensive and often limited by the coverage and quality of expert data. More recent human-in-the-loop methods use corrective interventions~\cite{peng2023learning,cai2025predictive,cai2025robot} or trajectory preferences~\cite{sadigh2017active,choi2020fast,knox2022models,cai2025predictive,wang2022feedback} to refine policies beyond demonstrations.










\vspace{-2mm}
\section{Method}
\label{Sec:method}
\vspace{-2mm}

In this section, we introduce~\ModelName, a human-preference
flow policy for long-horizon sidewalk navigation. We first present the
preliminaries in Sec.~\ref{Subsec:pre-method} and then describe the model architecture for imitation pretraining and the preference learning algorithm for alignment in Sec.~\ref{Subsec:model}.

\begin{figure}[t!] 
\centering
\includegraphics[width=\columnwidth]{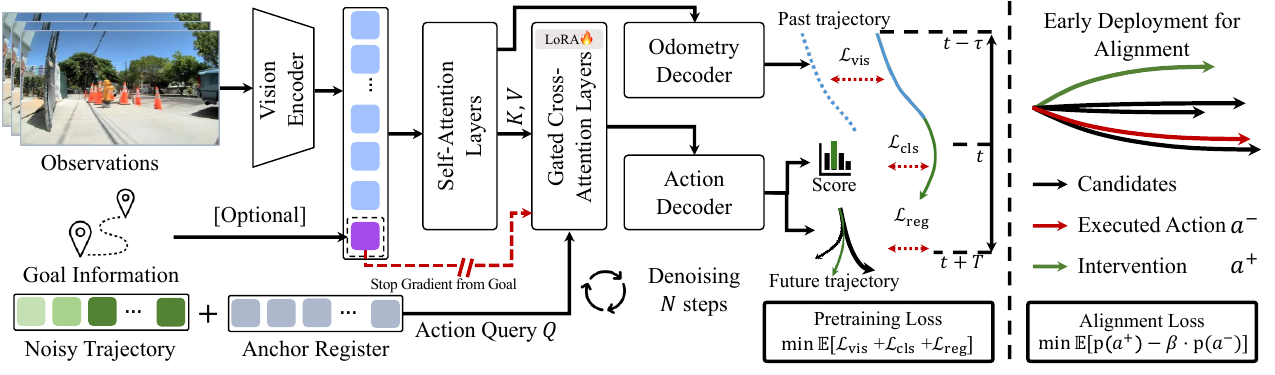} 
\caption{\textbf{Architecture and training pipeline of \ModelName.}
Given RGB observations and an optional goal, the encoder first extracts scene tokens, which are processed by self-attention layers and then used as keys and values in gated cross-attention.
A set of noisy trajectory queries, together with learnable anchor registers, is denoised to generate multiple trajectory candidates. 
During pretraining, the model is supervised with visual odometry prediction, mode classification, and trajectory regression losses. 
During deployment, human interventions are collected to align the policy by increasing the likelihood of corrected actions $a^{+}$ and suppressing undesirable executed actions $a^{-}$.
}
\label{fig:Method}
\vspace{-1em}
\end{figure}

\vspace{-1mm}
\subsection{Preliminaries}
\label{Subsec:pre-method}
\vspace{-1mm}

\paragraph{Task setting} We focus on \textbf{mapless} GPS-guided sidewalk navigation from a \textbf{monocular RGB camera}, one of the most challenging navigation settings. The objective of the robot is to reach a destination while maintaining safety and social compliance, \ie, avoiding collisions with pedestrians and obstacles, staying on the sidewalk, and minimizing contact with the grass. At each timestep $t$, the robot receives continuous RGB
frames $o_{t-T:t}$ from the past $T$ steps, a GPS goal $g_t \in \mathcal{G}$, and its pose $p_t$. The policy $\mathcal{\pi}_{\Theta}$ needs to plan a trajectory $\mathcal{P}=\left\{p_{t+1},...,p_{t+T} \right\}$ to generate control commands. Both $p_t$ and $g_t$ are highly noisy, making goal conditioning and long-horizon planning particularly challenging and leading to brittle behaviors if the policy over-relies on them. 

\vspace{-1mm}
\paragraph{Conditional flow matching policy} 
Given a data sample $(x^1, s_t)\sim\mathcal{D}$, where $x^1\sim\mathcal{P}$ denotes the planned future trajectory and $s_t=\{o_{t-T:t}, g_t, p_t\}$ denotes the current navigation state, we sample an initial noisy trajectory $x^0\sim\mathcal{N}(0,\mathbf{I})$, then adopt Rectified Flow~\cite{liu2022flow} to generate 
\begin{equation}
    x^{\tau} = (1-\tau)\cdot x^0 + \tau\cdot x^1, \quad \tau \sim \mathcal{U}(0,1),
    \label{eq:plan_interpolation}
\end{equation}
We train a velocity model $\boldsymbol{v}_{\Theta}(x^{\tau},s_t,\tau)$ to regress the target velocity field: \begin{equation} \mathcal{L}_{\text{FM}} = \mathbb{E}_{(x^1,s_t)\sim\mathcal{D},\,x^0\sim\mathcal{N}(0,\mathbf{I}),\,\tau\sim\mathcal{U}(0,1)} \left[ \left\| \boldsymbol{v}_{\Theta}(x^{\tau},s_t,\tau) - \boldsymbol{v} \right\|^2 \right], \label{eq:fm_policy} \end{equation} where the target velocity is $\boldsymbol{v}=x^1-x^0$. During inference, we initialize $x^0\sim\mathcal{N}(0,\mathbf{I})$ and solve the learned flow ODE  to obtain the predicted trajectory $\hat{x}^1
=
x^0+
\int_{0}^{1}
\boldsymbol{v}_{\Theta}(x^{\tau},s_t,\tau)
\,d\tau$.

\vspace{-1mm}
\paragraph{Preference learning} Recent works of preference-based reinforcement learning have explored the use of preference datasets to align policy with human expectations. Given a dataset $\mathcal{D}_{\text{pref}}$ consisted of preference pairs $(o_{t-T:t},g_t,p_t;x^+,x^-)\sim \mathcal{D}_{\text{pref}}$ means that the expert prefers the action $x^+$ over $x^-$ given state $s_t=\left\{o_{t-T:t},g_t,p_t \right\}$. Then we can train the policy $\pi_{\Theta}$ using the following objective~\cite{rafailov2023direct}
\begin{equation}
    \mathcal{L}_{\text{Pref}}=-\mathbb{E}[\log\sigma(\beta\cdot (\log\pi_{\Theta}(x^+|s_t)-\log\pi_{\Theta}(x^-|s_t)))],
    \label{eq:pref}
\end{equation}
where \noindent{$\sigma(\cdot)$} is the Sigmoid function, and $\beta > 0$ is a hyperparameter.

\vspace{-1mm}
\subsection{\ModelName\ Framework}
\label{Subsec:model}
\vspace{-1mm}

\begin{wrapfigure}{r}{0.52\textwidth}
\vspace{-5mm}
\centering
\includegraphics[width=\linewidth]{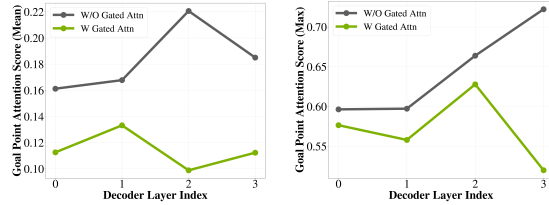}
\vspace{-4mm}
\caption{\textbf{Mitigating attention sink with gated attention.} 
The proportion of attention allocated to the \emph{goal / anchor} across layers. 
The baseline exhibits attention-sink behavior, in which a large fraction of attention is concentrated on the goal, whereas gated attention substantially reduces this concentration and encourages better context utilization.}
\label{fig:Attn}
\vspace{-4mm}
\end{wrapfigure}

\paragraph{Model architecture} As illustrated in Fig.~\ref{fig:Method}, we first introduce the anchor design. Given state-action pair $(s_t,x^1)\sim \mathcal{D}$ sampled from demonstration dataset $\mathcal{D}$, we parameterize each trajectory $x^1$ in the local frame, then normalize the action space to $[-1,1]^{T\times2}$. Then we apply a constrained $K$-means clustering over the normalized action space
to obtain a set of $K$ anchor actions $\{\bar x_k\}_{k=1}^{K}$ as prototypical multimodal behaviors for subsequent mode scoring and trajectory refinement. 
During pretraining, each expert trajectory $x^1$ is assigned to its nearest anchor,
$h=\arg\min_{k}\|x^1-\bar x_k\|_2^2$. We sample noise $x^0\sim \mathcal{N}(0,\boldsymbol{I})$ and a flow timestep $\tau \sim \text{Beta}(\alpha=1.5,\beta=1.0)$~\cite{bjorck2025gr00t}. The intermediate trajectory is constructed by linearly interpolating between noise and the expert trajectory,
$\bar x_h^{\tau}=(1-\tau)\cdot x^0+\tau\cdot \bar x_h$, with the target velocity field defined as $\boldsymbol{v}=x^1-x^0$. Conditioned on $(s_t, \tau,\bar x_h)$, the model predicts a velocity field $\boldsymbol{v}_{\Theta}(x^\tau,s_t,\tau,\bar x_h)$ and regresses it using an L2 loss. In parallel, a  scoring head $h_{\Theta}(s_t)$ predicts the anchor assignment using cross-entropy loss $\text{CE}\big(h_{\Theta}(s_t), h\big)$.

\begin{figure}[t!] 
\centering
\includegraphics[width=\columnwidth]{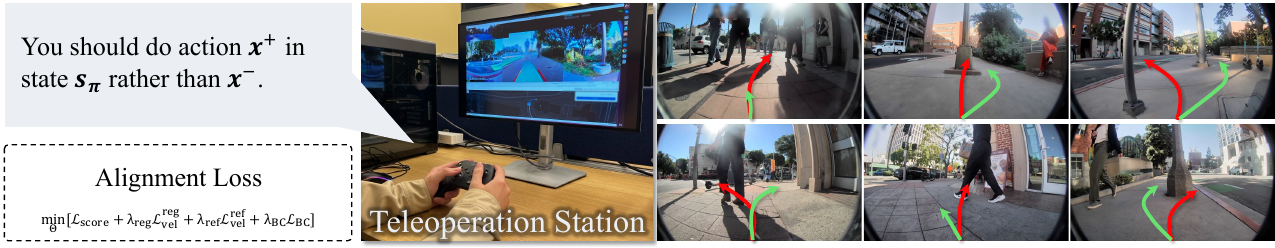} 
\caption{\textbf{Preference alignment with human intervention for a specific robot.}
Human monitors policy predictions and intervenes when inappropriate behaviors occur. The \textcolor{lightgreen}{corrective action $x^+$} is treated as preferred over the \textcolor{lightred}{policy action $x^-$}, providing pairwise feedback for alignment.
}
\label{fig:Preference}
\vspace{-1em}
\end{figure}

To enable context-aware, goal-conditioned learning, we design the architecture shown in Fig.~\ref{fig:Method}. We encode the observation history $o_{t-T:t}$ into features $V_{t-T:t}=[V_{t-T:t-1};V_t]$ using an ImageNet-pretrained\cite{deng2009imagenet} FastViT\cite{vasu2023fastvit}, and encode the goal $g_t$ into $G_t$. Self-attention is applied over $V_{t-T:t}$ to obtain $\tilde V_{t-T:t}=\text{MSA}(V_{t-T:t})$. We also add two auxiliary heads that predict the frame-to-frame displacement from $\tilde V_{t-T:t-1}$ and the future trajectory, supervised with MSE losses $\mathcal{L}_{\text{vis}}$.

Given the context $C_t=[\tilde V_{t-T:t};G_t]$, a noised query embedding 
$\boldsymbol{q}^{\tau}=\text{MLP}([\bar x_h^{\tau};\tau])$ attends to the context through gated cross-attention
$
\boldsymbol{z}^{\tau} = 
\text{softmax}\!\left(
\frac{\boldsymbol{q}^{\tau}\cdot \sigma(C_t W_K)^\top}{\sqrt{d_h}}
\right)\cdot \sigma(C_t W_V),
$
where $\sigma(\cdot)$ is the sigmoid function and $W_K,W_V$ are learnable projections. The latent $\boldsymbol{z}^{\tau}$ is then decoded to predict the velocity. As shown in Fig.~\ref{fig:Attn}, this gated attention mitigates attention sink on the goal and encourages the policy to use scene cues such as obstacles and sidewalk boundaries.

\vspace{-1mm}
\paragraph{Preference alignment with  human intervention} Pure imitation learning remains limited by the coverage of demonstrations: the policy may fail in unseen safety-critical states, struggle to recover from distributional shift, or produce behaviors that are socially non-compliant. More importantly, although large-scale pretraining yields a broadly capable navigation policy, it does not necessarily provide the precise, robot-specific control. To bridge this gap, we further adapt the pretrained policy from broad demonstration-driven competence to robot-specific behavioral precision through human-in-the-loop preference alignment. 
As illustrated in Fig.~\ref{fig:Preference}, we deploy the policy $\pi_{\Theta}$ in the real world and allow a human to monitor the trajectory through the teleoperation platform. Whenever there is an unsafe or socially inappropriate prediction, the operator intervenes and provides corrective commands. Such interventions naturally focus on critical states $s_{\pi}$ where pure imitation is insufficient. In each intervention, the human corrective action $x^+$ is treated as the preferred action, while the policy action $x^-=\pi_{\Theta}(s_{\pi})$ is treated as the less preferred action. This yields a set of pairwise preference samples $(s_{\pi},x^+,x^-)$ that capture specific requirements of safety and social compliance.

Based on Eq.~\ref{eq:pref}, we optimize the policy with preference objectives and regularize~\cite{liu2025flow, liu2025improving} it toward a reference policy, \ie, the pretrained policy $\pi_{\Theta_0}=(\boldsymbol{v}_{\Theta_0},h_{\Theta_0})$. Given a preference pair $(x^+, x^-)$, where $x^+$ denotes the human-preferred corrective action and $x^-$ denotes the less-preferred policy prediction, we assign their corresponding anchor modes as $h^+$ and $h^-$. The scoring head is then encouraged to assign higher probability to the preferred anchor than to the rejected anchor. We use the score-level preference margin
$
    \Delta_{\Theta}^{\text{score}}
    =
    \log p_{\Theta}(h^+|s_t)
    -
    \log p_{\Theta}(h^-|s_t),
$ and $\mathcal{L}_{\text{score}}=-\log(\sigma(\beta\cdot\Delta_{\Theta}^{\text{score}}))$ for finetuning.
To preserve the prior, we regularize the scoring head toward the original distribution from the reference policy. Specifically, we define
$
    p_{\Theta}(k|s_t)
    =
    \mathrm{softmax}\left(h_{\Theta}(s_t)\right)_k
$ and $p_{\Theta_0}(k|s_t)
    =
    \mathrm{softmax}\left(h_{\Theta_0}(s_t)\right)_k$
. We then use a KL regularization term:
\begin{equation}
    \mathcal{L}_{\mathrm{score}}^{\mathrm{reg}}
    =
    D_{\mathrm{KL}}
    \left(
    p_{\Theta}(\cdot|s_t)
    \,\|\, 
    p_{\Theta_0}(\cdot|s_t)
    \right).
\end{equation}
For the velocity estimation network, unlike previous works that explicitly reduce the likelihood of rejected actions~\cite{liu2025flow, liu2025improving}, we do not directly decrease the probability of the rejected action $x^-$. Since the velocity field models a continuous action distribution, directly pushing away $x^-$ may introduce unstable gradients and damage nearby feasible actions. Instead, we use an asymmetric trajectory-level objective: the preferred corrective action is used as a positive regression target, while the rejected mode is regularized toward its anchor prior rather than toward the rejected policy prediction:
\begin{equation}
\begin{aligned}
    \mathcal{L}_{\text{vel}}
    =
    &\;
    \mathbb{E}_{\tau,\epsilon}
    \left[
    \left\|
    \boldsymbol{v}_{\Theta}(s_t, \bar x_{h^+}^{\tau}, \tau, h^+)
    -
    v^+
    \right\|^2
    \right]
    +
    \lambda_{\text{anc}}
    \mathbb{E}_{\tau,\epsilon}
    \left[
    \left\|
    \boldsymbol{v}_{\Theta}(s_t, \bar x_{h^-}^{\tau}, \tau, h^-)
    -
    v^-
    \right\|^2
    \right],
\end{aligned}
\end{equation}
where $v^+=x^+-x^0,v^-=\bar x_{h-}-x^0$. And to avoid overfitting to sparse corrective feedback and preserve the prior, we further regularize the finetuned policy toward the reference policy $\pi_{\Theta_0}$ using:
\begin{equation}
\begin{aligned}
    \mathcal{L}_{\text{vel}}^{\text{reg}}
    =
    \mathbb{E}_{\tau,\epsilon,h=(h^+,h^-)}
    \left[
    \left\|
    \boldsymbol{v}_{\Theta}(s_t, \bar x_{h}^{\tau}, \tau, h)
    -
    \operatorname{sg}\left[
    \boldsymbol{v}_{\Theta_0}(s_t, \bar{h}_{\tau}^{h}, \tau, h)
    \right]
    \right\|^2
    \right],
\end{aligned}
\end{equation}
where $\operatorname{sg}[\cdot]$ denotes the stop-gradient operation. Together, the preference alignment objective is
\begin{equation}
    \mathcal{L}_{\text{Pref}}
    =
    \mathcal{L}_{\text{score}}
    +
    \lambda_{\text{reg,s}}\mathcal{L}_{\mathrm{score}}^{\mathrm{reg}}
    +
    \lambda_{\text{v}}
    \mathcal{L}_{\text{vel}}
    +
    \lambda_{\text{reg,v}}
    \mathcal{L}_{\text{vel}}^{\text{reg}}
    +
    \lambda_{\text{FM}}
    \mathcal{L}_{\text{FM}},
\end{equation}
where $\mathcal{L}_{\text{FM}}$ is the original flow matching loss.
\vspace{-2mm}
\section{Experiments}
\label{sec:experiments}
\vspace{-2mm}

\begin{figure}[t]
\vspace{-2mm}
\centering
\begin{minipage}[t]{0.36\linewidth}
    \vspace{0pt}
    \centering
    \includegraphics[width=\linewidth]{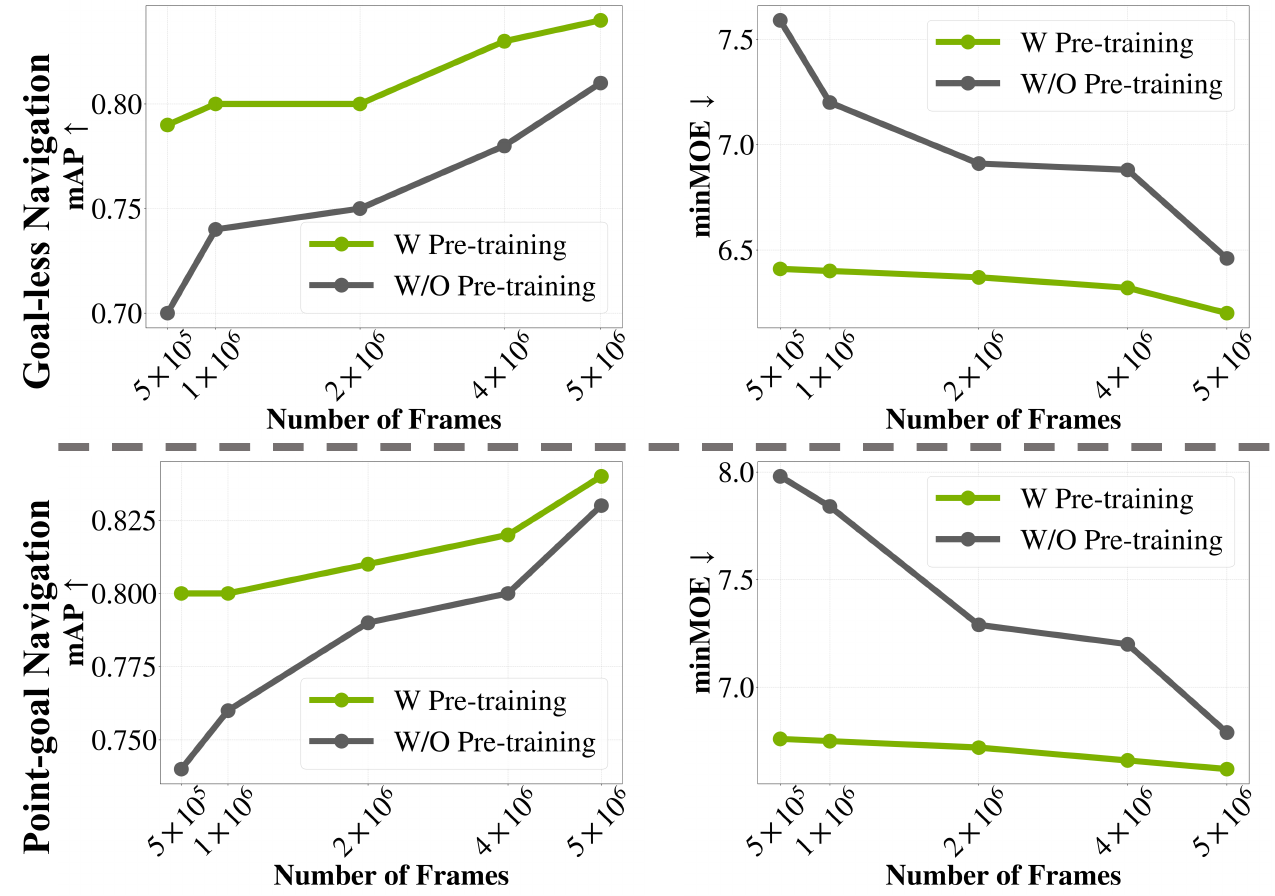}
    \vspace{-4mm}
    \captionof{figure}{\textbf{Effectiveness of pretraining on mixed dataset.}}
    \label{fig:u-pretraining}
\end{minipage}
\hspace{0.02\linewidth}
\begin{minipage}[t]{0.6\linewidth}
    \vspace{0pt}
    \centering
    \footnotesize
    \resizebox{\linewidth}{!}{%
    \begin{tabular}{lcccc}
    \toprule
    \textbf{Method} & minMOE~$\downarrow$ & minADE~$\downarrow$ & L2~$\downarrow$ & mAP~$\uparrow$ \\
    \midrule
    GNM~\cite{shah2022gnm} & 7.31 & 0.63 & 1.32 & 0.63 \\
    ViNT~\cite{shah2023vint} & 8.51 & 0.74 & 1.54 & 0.74 \\
    NoMaD~\cite{sridhar2024nomad} & 13.77 & 1.34 & 2.71 & 0.63 \\
    CityWalker~\cite{liu2024citywalker} & 8.94 & 0.71 & 1.48 & 0.77 \\
    MIMIC~\cite{he2026learning} & 9.31 & 0.52 & 1.43 & 0.69 \\
    S2E~\cite{he2025seeing} & 6.77 & \textbf{0.46} & 1.73 & 0.81 \\
    DiffusionDrive~\cite{liao2025diffusiondrive} & 6.96 & 0.63 & 1.71 & 0.77 \\
    \midrule
    \ours \ModelNameIL
    & \textbf{6.63} & 0.49 & \textbf{1.04} & \textbf{0.87} \\
    \bottomrule
    \end{tabular}%
    }
    \captionof{table}{\textbf{Open-loop benchmark results.}
    All models are retrained on the same dataset for a fair comparison.}
    \label{tab:open-loop}
\end{minipage}
\vspace{-3mm}
\end{figure}

\subsection{Implementation Details}
\label{sub:impl}

\vspace{-1mm}
\paragraph{Dataset}
For pretraining, we collect \textbf{\Dataset}, a large-scale 300-hour video teleoperation dataset with 48,508 trajectory segments and about 19.4 million frames. The data are collected by a fleet of 281 robot instances navigating sidewalks across different cities, teleoperated by humans. Each segment contains 1920$\times$1080 fisheye RGB video at 20 Hz, synchronized with robot states. For finetuning, we collect about 13~K frames of real-world intervention data and simulation intervention data generated by an A$^\ast$~\cite{hart1968formal} planner.  More details are provided in the \underline{Appendix}.

\vspace{-1mm}
\paragraph{Training details}
The model takes 11 consecutive frames as input and predicts a 4-second trajectory at 20 Hz. For real-time deployment and fair comparison, all methods use fixed 352$\times$128 inputs and the same frame rate. 
We train the model in three stages. First, we pretrain the backbone policy on a mixture of large-scale video-action datasets~\cite{zhang2024toward, akhtyamov2025egowalk, liang2025gnd, hirose2023sacson, nguyen2023toward, hirose2025learning, karnan2022socially, dauner2024navsim, nvidia2025physicalaiav} collected from different robots and environments. To handle heterogeneous distributions, we append an embodiment ID token~\cite{bjorck2025gr00t} to the visual-temporal tokens and apply a dataset-specific normalization factor to the target trajectories. After pretraining, we perform supervised finetuning (SFT) on our dataset to adapt the policy to sidewalk navigation. This stage uses the same observation and trajectory prediction format as pretraining, but focuses on real-world sidewalk behaviors such as lane following, obstacle avoidance, pedestrian interaction, curb handling, and long-horizon goal-conditioned navigation.
Finally, for preference alignment, we freeze all weights except the LoRA~\cite{hu2022lora} modules applied to gated cross attention layers. 
More details are in the \underline{Appendix}.

\vspace{-1mm}
\paragraph{Baselines}
We compare with state-of-the-art visual navigation models, including GNM~\citep{shah2023gnm}, ViNT~\citep{shah2023vint}, NoMaD~\citep{sridhar2024nomad}, MBRA~\citep{hirose2025learning}, CityWalker~\citep{liu2024citywalker}, MIMIC~\citep{he2026learning}, and S2E~\citep{he2025seeing}. 
We also compare different policy adaptation strategies~\citep{kelly2019hg,menda2019ensembledagger,seneviratne2025halo} using the same intervention data on the simulator~\citep{mittal2025isaac,wu2025towards}. 
All baselines are \textbf{retrained} with the same protocol and data splits for a fair comparison. 

\vspace{-1mm}
\subsection{Open-loop Evaluation}
\label{exp:openloop}
\vspace{-1mm}

Table~\ref{tab:open-loop} summarizes the open-loop performance of different visual navigation models on the held-out evaluation set. We report minMOE~\cite{liu2024citywalker,chen2025socialnav} to evaluate the minimum error between the generated trajectories and the ground-truth trajectory, while minADE and L2 measure the trajectory prediction accuracy in terms of average and endpoint distance errors. We further use mean average precision (mAP) to evaluate whether the model assigns higher confidence to accurate trajectory modes~\cite{ettinger2021large}. Compared with prior methods, \ModelName\ achieves the best performance on minMOE, L2, and mAP, while maintaining competitive minADE. These results show that the proposed architecture not only predicts more accurate future trajectories, but also ranks its multi-modal predictions more reliably.

As shown in Fig.~\ref{fig:u-pretraining}, pretraining on mixed dataset consistently improves performance across different data scales and task settings. With more training frames, both the pretrained and non-pretrained models improve, but the pretrained model achieves better mAP and lower minMOE throughout the scaling curve. This suggests that unified pretraining provides a strong robot-agnostic navigation prior, enabling the model to leverage downstream sidewalk data more effectively.

\vspace{-1mm}
\subsection{Closed-loop Evaluation}
\label{exp:closed}
\vspace{-1mm}

\begin{figure*}[t]
    \centering
    \begin{minipage}[t]{0.49\textwidth}
        \vspace{0pt}
        \centering
        \includegraphics[width=\linewidth]{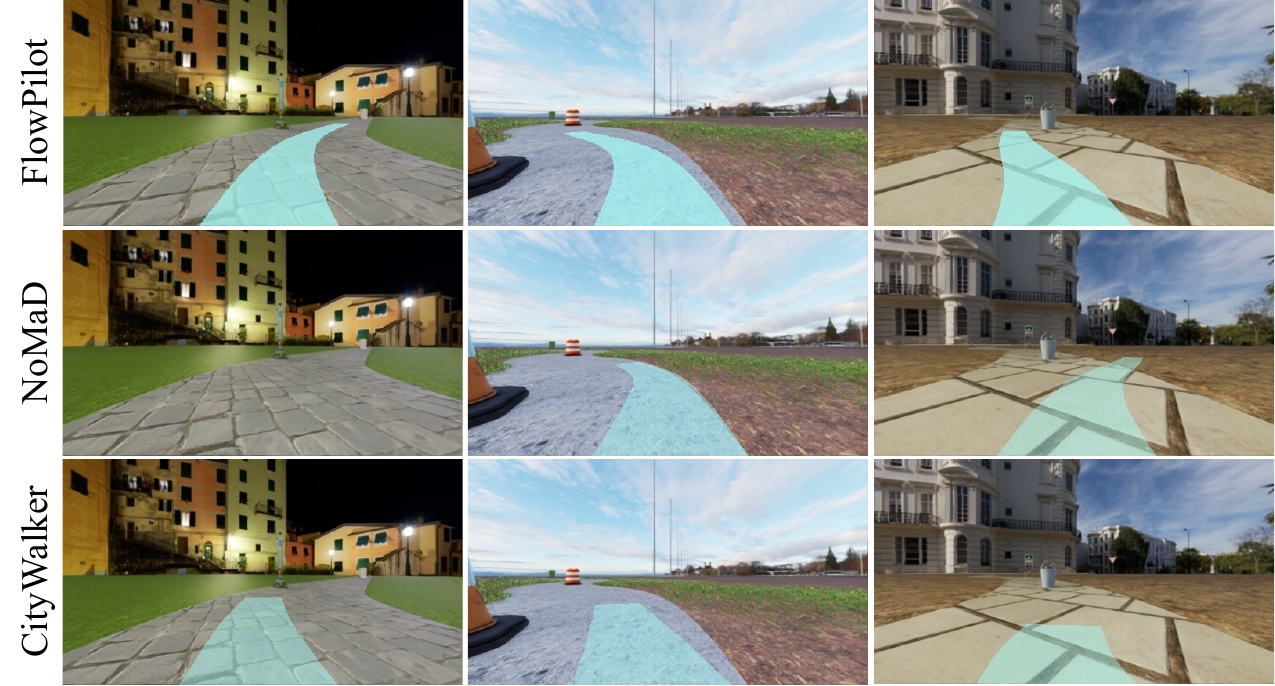}
        \caption{\textbf{Qualitative comparison in simulated sidewalk environments.}}
        \label{fig:sim-qualitative}
    \end{minipage}
    \hspace{0.03\textwidth}
    \begin{minipage}[t]{0.45\textwidth}
        \vspace{0pt}
        \centering
        {\footnotesize
        \setlength{\tabcolsep}{3.5pt}
        \resizebox{\linewidth}{!}{
        \begin{tabular}{lcccc}
        \toprule
        \textbf{Method} 
        & SR~$\uparrow$ 
        & RC~$\uparrow$ 
        & CR~$\downarrow$ 
        & OBR~$\downarrow$ \\
        \midrule
        
        GNM~\citep{shah2023gnm}
        & 0.19 &0.39 &0.68 & \textbf{0.01}\\
        
        ViNT~\cite{shah2023vint}
        & 0.27 & 0.51 & 0.60 & 0.04 \\
        
        NoMaD~\cite{sridhar2024nomad}
        & 0.21 & 0.45 & 0.74 & 0.04\\
        
        MBRA~\cite{hirose2025learning}
        & 0.31 & 0.50 & 0.54 & 0.15\\
        
        CityWalker~\cite{liu2025citywalker}
        & 0.28 & 0.48 & 0.70 &  0.02\\
        
        MIMIC~\cite{he2026learning}
        & 0.25 & 0.44 & 0.62 & 0.03\\
        
        S2E~\cite{he2025seeing}
        & 0.35 & 0.55 & 0.51 & 0.13\\

        \ModelName\ w/o GCA &0.29 &0.59 &0.61 &0.09 \\
        \ours\ModelNameIL\ & \textbf{0.42} & \textbf{0.66} & \textbf{0.43} & \textbf{0.01} \\
        
        \bottomrule
        \end{tabular}}}
    \vspace{-1.5mm}
    \captionof{table}{\textbf{Closed-loop evaluation with robot R0 on the simulation benchmark.}}
        \label{tab:sim-closed-loop}
    \end{minipage}
\end{figure*}

\paragraph{Unit-test benchmarks}
We evaluate different navigation models and conduct ablation studies on a closed-loop simulation benchmark built upon Urban-Sim~\citep{wu2025towards,mittal2025isaac}. We first compare navigation methods on the basic robot configuration R0, which matches the average dynamics and camera of the wheeled robots in our training data. We then evaluate human-in-the-loop learning on R0 and two novel configurations, R1 and R2, with shifted dynamics and cameras such as a 20\% change in focal length. This setting tests adaptation to deployment-specific embodiment shifts. Each scenario contains a 20-meter sidewalk segment with obstacles, and each method is evaluated over 300 trials. We report success rate (SR), route completion (RC), collision rate (CR), and off-boundary rate (OBR), where collisions, off-boundary behaviors, and timeouts are counted as failures. Intervention data $\mathcal{D}_{\text{Int}}$ are collected from the same scenario distribution but are disjoint from the test scenarios.

\begin{table*}[t]
\centering
\scriptsize
\setlength{\tabcolsep}{2.4pt}
\resizebox{\textwidth}{!}{
\begin{tabular}{l*{15}{c}}
\toprule
& \multicolumn{5}{c}{\textbf{Basic Robot Configuration R0}}
& \multicolumn{5}{c}{\textbf{Novel Robot Configuration R1}}
& \multicolumn{5}{c}{\textbf{Novel Robot Configuration R2}} \\
\cmidrule(lr){2-6} \cmidrule(lr){7-11} \cmidrule(lr){12-16}
\textbf{Method} 
& $\mathcal{D}_{\text{Int}}$ & SR~$\uparrow$ & RC~$\uparrow$ & CR~$\downarrow$ & OBR~$\downarrow$
& $\mathcal{D}_{\text{Int}}$ & SR~$\uparrow$ & RC~$\uparrow$ & CR~$\downarrow$ & OBR~$\downarrow$
& $\mathcal{D}_{\text{Int}}$ & SR~$\uparrow$ & RC~$\uparrow$ & CR~$\downarrow$ & OBR~$\downarrow$\\
\midrule

\ModelNameIL
& -- & {0.42} & {0.66} & {0.43} & \textbf{0.01}
& -- & {0.29} & {0.52} & 0.64 & 0.07
& -- & 0.24 & 0.42 & 0.21 & 0.01 \\

\midrule
BC
& 265~K & 0.22 & 0.44 & 0.28 & 0.46
& 265~K & 0.27 & 0.51 & 0.71 & 0.02
& 265~K & 0.28 & 0.49 & 0.35 & 0.37 \\

HG-DAgger~\cite{kelly2019hg}
& 46~K & 0.31 & 0.51 & 0.24 & 0.30
& 50~K & 0.27 & 0.48 & 0.43 & 0.30
& 91~K & 0.23 & 0.43 & 0.29 & 0.48 \\

Ensemble-DAgger~\cite{menda2019ensembledagger}
& 60~K & 0.14 & 0.38 & \textbf{0.13} & 0.73 
& 77~K & 0.25 & 0.46 & 0.30 & 0.45
& 80~K & 0.22 & 0.42 & 0.33 & 0.45 \\

\ModelName\ w/o Ref.
& 46~K & 0.29 & 0.47 & 0.50 & 0.18
& 50~K & 0.21 & 0.43 & 0.36 & 0.39
& 91~K & 0.26 & 0.50 & 0.20 & 0.47 \\

\ours\ModelNameFull
& 46~K & \textbf{0.55} & \textbf{0.74} & 0.31 & {0.14}
& 50~K & \textbf{0.44} & \textbf{0.66} & 0.43 & 0.13
& 91~K & \textbf{0.35} & \textbf{0.63} & 0.49 & 0.16 \\

\bottomrule
\end{tabular}
}
\caption{\textbf{Closed-loop evaluation on the simulation benchmark with different robots.} 
}
\vspace{-1em}
\label{tab:robotconfig}
\end{table*}

\begin{table*}[t]
\centering
\scriptsize
\setlength{\tabcolsep}{7pt}
\resizebox{\textwidth}{!}{
\begin{tabular}{lccccccc}
\toprule
\textbf{Metric} 
& ViNT~\cite{shah2023vint}
& NoMaD~\cite{sridhar2024nomad}
& MBRA~\cite{hirose2025learning}
& CityWalker~\cite{liu2025citywalker}
& MIMIC~\cite{he2026learning}
& S2E~\cite{he2025seeing}
& \ModelName \\
\midrule
NIR~$\downarrow$
& 0.212 & 0.142 & 0.103 & 0.055 & 0.061 & 0.054 & \textbf{0.034} \\
NCR~$\downarrow$
& 0.043 & 0.040 & 0.044 &0.035 & 0.030 & \textbf{0.021} & 0.023 \\
NOBR~$\downarrow$
& 0.169 & 0.101 & 0.059 & 0.019 & \textbf{0.009} & 0.033 & 0.011 \\
\bottomrule
\end{tabular}}
\caption{\textbf{Closed-loop long-horizon evaluation on the simulation benchmark.} 
}
\label{tab:longhorizon}
\end{table*}

\vspace{-2mm}
\paragraph{Long-horizon benchmarks} We further evaluate long-horizon navigation performance to evaluate deployment-scale robustness over extended routes. Each episode uses start and goal locations separated by more than 100~m. When a failure occurs, we do not terminate the episode; instead, we reset the robot to the closest valid waypoint along the route, continue the route, and record the event as an intervention. The benchmark contains 100 procedurally generated scenes and covers 36~km of total route distance. We report Normalized Intervention Rate (NIR)~\cite{zhang2025creste}, Normalized Collision Rate (NCR), and Normalized Out-of-Boundary Rate (NOBR), which measure intervention events, collision events, and sidewalk-boundary violations normalized by traveled distance, respectively.


\vspace{-2mm}
\paragraph{Results} As shown in Tab.~\ref{tab:sim-closed-loop} and Fig.~\ref{fig:sim-qualitative}, \ModelNameIL\ performs the best across all metrics, demonstrating the effectiveness of the proposed method. The ablation further shows the benefit of Gated Cross Attention (GCA); compared with the one without GCA, \ModelNameIL\ improves SR from 0.29 to 0.42 and reduces CR from 0.61 to 0.43, highlighting the importance of GCA for robust context understanding for navigation. For cross-robot adaptation, Tab.~\ref{tab:robotconfig} shows that the model trained with human preference \ModelNameFull\ achieves the highest SR and RC across R0, R1, and R2, demonstrating effective adaptation to shifted robot dynamics and camera configurations. Compared with the one without reference policy regularization (\ModelName\ w/o Ref.), \ModelNameFull\ consistently improves both SR and RC, indicating that reference regularization helps preserve the pretrained navigation prior during finetuning. And for the long-horizon task, Tab.~\ref{tab:longhorizon} shows that \ModelName\ achieves the lowest NIR, outperforming all baselines in large-scale scenarios.

\vspace{-1mm}
\subsection{Real-world Deployment}
\label{exp:Real}
\vspace{-1mm}

\begin{figure*}[t]
    \centering
    \begin{minipage}[t]{0.46\textwidth}
        \vspace{0pt}
        \centering
        \includegraphics[width=\linewidth]{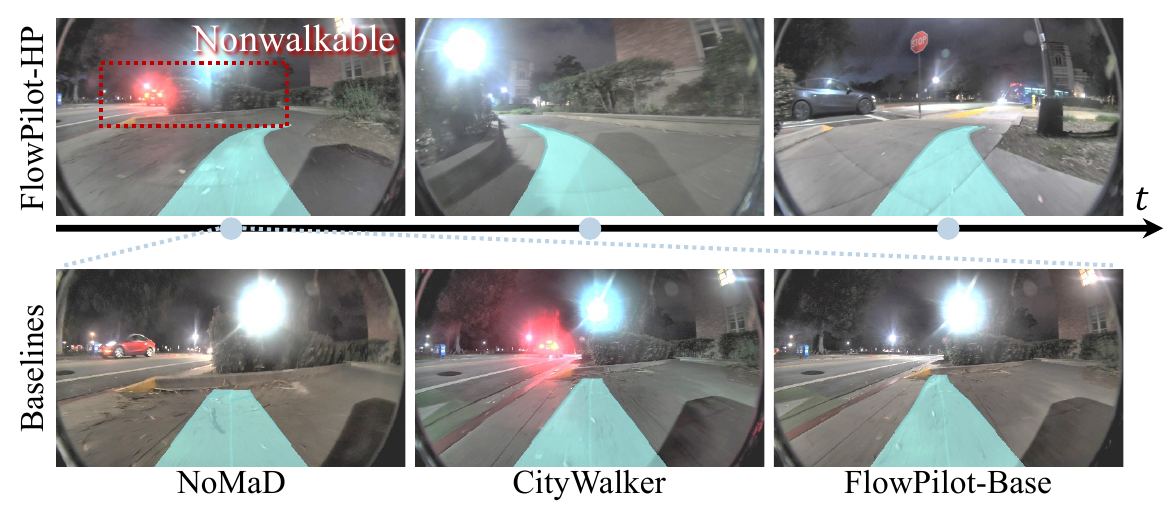}
        \vspace{-6mm}
        \caption{\textbf{Qualitative comparison in real-world sidewalk environments.}}
        \label{fig:real-world-comparison}
    \end{minipage}
    \hspace{0.02\textwidth}
    \begin{minipage}[t]{0.50\textwidth}
        \vspace{2pt}
        \centering
        {\footnotesize
        \setlength{\tabcolsep}{3.5pt}
        \resizebox{\linewidth}{!}{
        \begin{tabular}{l cccc}
            \toprule
            Method 
            & IR~$\downarrow$ 
            & NIR~$\downarrow$ 
            & DCR~$\uparrow$ 
            & TCR~$\uparrow$ \\
            \midrule
            ViNT~\cite{shah2023vint}
            & 0.081 & 0.304 & 0.867 & 0.858 \\
            NoMaD~\cite{sridhar2024nomad}
            & 0.093 & 0.423 & 0.885 & 0.864 \\
            CityWalker~\cite{liu2024citywalker}
            & 0.146 & 0.483 & 0.809 & 0.804 \\
            \ours \ModelNameIL
            & 0.020 & 0.073 & 0.904 & 0.916 \\
            \ours \ModelNameFull
            & \textbf{0.012} 
            & \textbf{0.035} 
            & \textbf{0.928} 
            & \textbf{0.939} \\
            \bottomrule
        \end{tabular}}}
        \vspace{-1mm}
        \captionof{table}{\textbf{Closed-loop evaluation in long-horizon real-world sidewalk environments.}}
        \label{tab:real-world-closed-loop}
    \end{minipage}
\end{figure*}

\begin{figure}[t!] 
\centering
\includegraphics[width=\columnwidth]{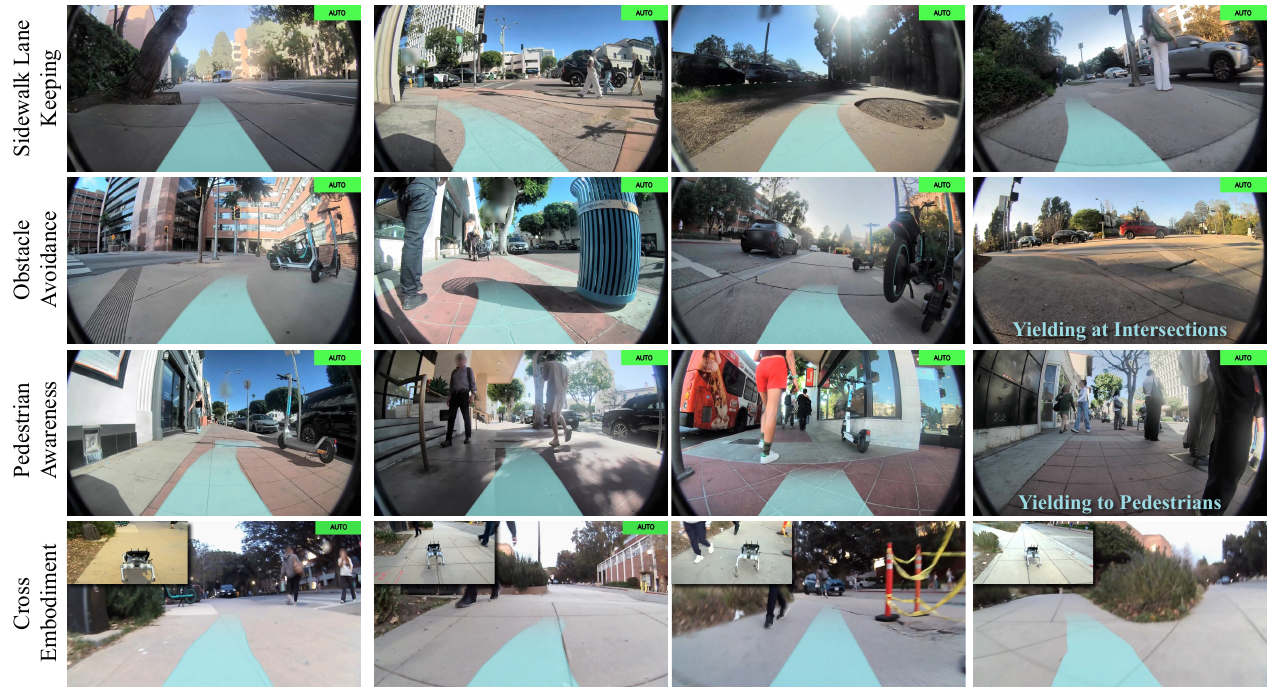} 
\caption{
\textbf{Qualitative results in diverse real-world sidewalk environments.}
We visualize the trajectories predicted by \ModelName\ and executed by the robot during real-world deployment.
}
\vspace{-4mm}
\label{fig:realworld}
\end{figure}

\vspace{-1mm}
\paragraph{Robot, scenarios and metrics}  To validate real-world navigation performance, we conduct experiments on a wheeled robot and a robot dog shown in Fig.~\ref{fig:teaser} and Fig.~\ref{fig:realworld} in 8 scenarios covering 16 routes, with a total route length of approximately 2.8~km. In real-world environments, a human operator takes over when the robot exhibits unsafe behavior, such as approaching obstacles or pedestrians, or drifting toward non-walkable regions. Since these interventions ensure safety, collision rate and success rate are less informative. We therefore report Intervention Rate (IR) and Normalized Intervention Rate (NIR)~\cite{zhang2025creste}. Specifically, IR is the fraction of distance driven under human takeover, \ie, $\mathrm{IR}=d_{\mathrm{intervention}}~/~d_{\mathrm{total}}$, while NIR measures the number of intervention events normalized by the traveled distance. We also report social-compliance metrics~\cite{chen2025socialnav}, including Distance Compliance Rate (DCR) and Time Compliance Rate (TCR), which measure the fraction of distance and time during which the robot remains socially compliant.

\begin{figure}[t!]
\centering
\includegraphics[width=\textwidth]{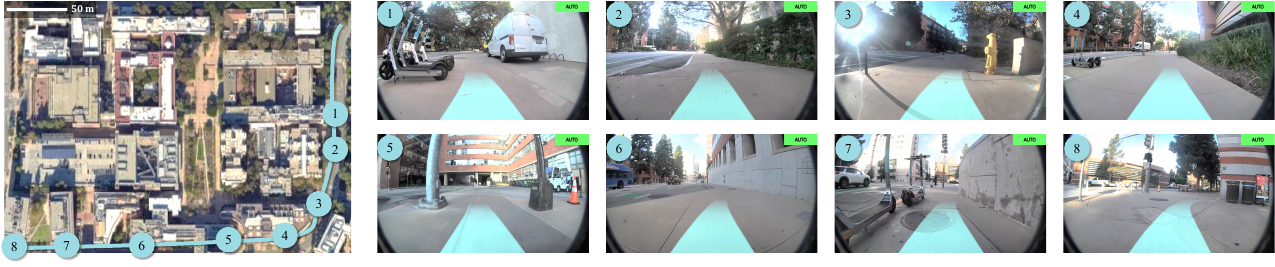}
\vspace{-6mm}
\caption{\textbf{Kilometer-scale long-horizon sidewalk navigation.} We evaluate \ModelName\ in a long-horizon, complex real-world scenario, demonstrating its robust progress over long distances.}
\label{fig:long}
\vspace{-2mm}
\end{figure}

\begin{figure}[t!]
\centering
\includegraphics[width=\textwidth]{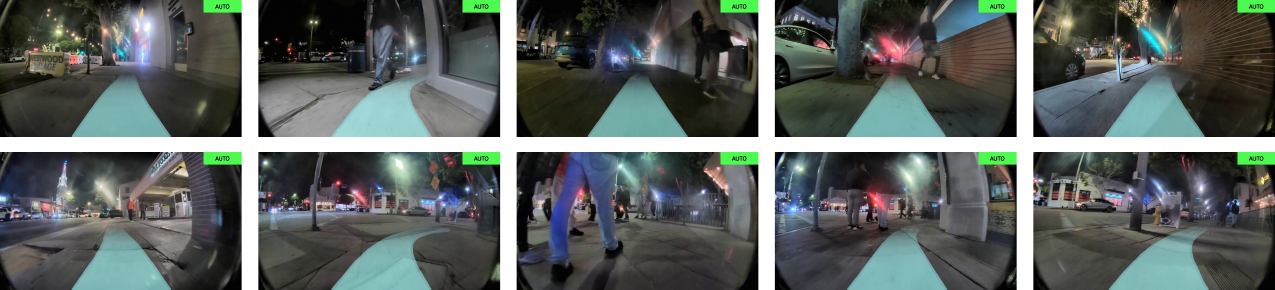}
\vspace{-4mm}
\caption{\textbf{Long-horizon sidewalk navigation under challenging nighttime illumination.}
We evaluate \ModelName\ on a long-horizon route in night with low-light regions and complex illumination changes, demonstrating its robustness to degraded visual conditions.}
\label{fig:long-2}
\vspace{-4mm}
\end{figure}

\vspace{-2mm}
\paragraph{Results}
As shown in Tab.~\ref{tab:real-world-closed-loop}, both \ModelName{}-Base and \ModelName{}-HP substantially outperform prior methods in real-world deployment, especially on intervention-related metrics. \ModelName{}-HP further reduces IR by 40.0\% and NIR by 52.1\% relative to \ModelName{}-Base, validating the effectiveness of human preference fine-tuning. Qualitatively, Fig.~\ref{fig:realworld}, Fig.~\ref{fig:long} and Fig.~\ref{fig:long-2} show that our model produces smooth and stable trajectories across diverse sidewalk scenarios, including sidewalk lane keeping, obstacle avoidance, pedestrian awareness, and cross-embodiment deployment. And as shown in Fig.~\ref{fig:real-world-comparison}, our method also exhibits more socially compliant behavior than baselines  where it maintains safer clearance and avoids collision.

\vspace{-2mm}
\section{Conclusion}
\label{sec:conclusion}
\vspace{-2mm}

We present {\ModelName}, a novel long-horizon sidewalk navigation framework. We introduce an anchored flow-matching policy with a gated transformer for long-horizon, GPS-guided mapless sidewalk navigation. It provides a powerful representation that enables the model to capture the complex features of sidewalk environments. Built on large-scale imitation learning, we then fine-tune the policy to transition from imitation to alignment, using paired preference data collected in real-world environments. Simulation and real-world experiments show that {\ModelName} outperforms baselines: \ModelNameIL\ achieves 42\% SR and 66\% RC in simulation, while \ModelNameFull\ improves robot-specific adaptation and real-world social compliance. In particular, human preference fine-tuning reduces IR by 40.0\% and NIR by 52.1\% relative to the base model.

\vspace{-2mm}
\section{Limitations} 
\vspace{-2mm}

Our current policy, based on a monocular RGB camera, is susceptible to scale ambiguity and occasionally struggles to understand fine-grained 3D structural cues. This can lead to performance degradation in scenarios involving textureless surfaces or reflective obstacles. Future iterations would benefit from integrating explicit depth estimation or occupancy prediction to provide the necessary geometric priors for safer navigation in dense, cluttered environments.

\clearpage

\bibliography{references}

@article{seneviratne2025halo,
  title={HALO: Human Preference Aligned Offline Reward Learning for Robot Navigation},
  author={Seneviratne, Gershom and An, Jianyu and Ellahy, Sahire and Weerakoon, Kasun and Elnoor, Mohamed Bashir and Kannan, Jonathan Deepak and Sunil, Amogha Thalihalla and Manocha, Dinesh},
  journal={arXiv preprint arXiv:2508.01539},
  year={2025}
}

@inproceedings{kelly2019hg,
  title={Hg-dagger: Interactive imitation learning with human experts},
  author={Kelly, Michael and Sidrane, Chelsea and Driggs-Campbell, Katherine and Kochenderfer, Mykel J},
  booktitle={2019 International Conference on Robotics and Automation (ICRA)},
  pages={8077--8083},
  year={2019},
  organization={IEEE}
}

@article{mittal2025isaac,
  title={Isaac lab: A gpu-accelerated simulation framework for multi-modal robot learning},
  author={Mittal, Mayank and Roth, Pascal and Tigue, James and Richard, Antoine and Zhang, Octi and Du, Peter and Serrano-Munoz, Antonio and Yao, Xinjie and Zurbr{\"u}gg, Ren{\'e} and Rudin, Nikita and others},
  journal={arXiv preprint arXiv:2511.04831},
  year={2025}
}

@article{akhtyamov2025egowalk,
  title={Egowalk: A multimodal dataset for robot navigation in the wild},
  author={Akhtyamov, Timur and Mdfaa, Mohamad Al and Benavides, Javier Antonio Ramirez and Nigmatzyanov, Arthur and Bakulin, Sergey and Devchich, German and Fatykhov, Denis and Salinas, Diego Ruiz and Mazurov, Alexander and Zipa, Kristina and others},
  journal={arXiv preprint arXiv:2505.21282},
  year={2025}
}

@inproceedings{liang2025gnd,
  title={Gnd: Global navigation dataset with multi-modal perception and multi-category traversability in outdoor campus environments},
  author={Liang, Jing and Das, Dibyendu and Song, Daeun and Shuvo, Md Nahid Hasan and Durrani, Mohammad and Taranath, Karthik and Penskiy, Ivan and Manocha, Dinesh and Xiao, Xuesu},
  booktitle={2025 IEEE International Conference on Robotics and Automation (ICRA)},
  pages={2383--2390},
  year={2025},
  organization={IEEE}
}

@article{hirose2025learning,
  title={Learning to drive anywhere with model-based reannotation},
  author={Hirose, Noriaki and Ignatova, Lydia and Stachowicz, Kyle and Glossop, Catherine and Levine, Sergey and Shah, Dhruv},
  journal={IEEE Robotics and Automation Letters},
  volume={11},
  number={2},
  pages={1242--1249},
  year={2025},
  publisher={IEEE}
}

@article{karnan2022socially,
  title={Socially compliant navigation dataset (scand): A large-scale dataset of demonstrations for social navigation},
  author={Karnan, Haresh and Nair, Anirudh and Xiao, Xuesu and Warnell, Garrett and Pirk, S{\"o}ren and Toshev, Alexander and Hart, Justin and Biswas, Joydeep and Stone, Peter},
  journal={IEEE Robotics and Automation Letters},
  volume={7},
  number={4},
  pages={11807--11814},
  year={2022},
  publisher={IEEE}
}

@inproceedings{nguyen2023toward,
  title={Toward human-like social robot navigation: A large-scale, multi-modal, social human navigation dataset},
  author={Nguyen, Duc M and Nazeri, Mohammad and Payandeh, Amirreza and Datar, Aniket and Xiao, Xuesu},
  booktitle={2023 IEEE/RSJ international conference on intelligent robots and systems (IROS)},
  pages={7442--7447},
  year={2023},
  organization={IEEE}
}

@article{hirose2023sacson,
  title={Sacson: Scalable autonomous control for social navigation},
  author={Hirose, Noriaki and Shah, Dhruv and Sridhar, Ajay and Levine, Sergey},
  journal={IEEE Robotics and Automation Letters},
  volume={9},
  number={1},
  pages={49--56},
  year={2023},
  publisher={IEEE}
}

@article{zhang2024toward,
  title={Toward robust robot 3-d perception in urban environments: The ut campus object dataset},
  author={Zhang, Arthur and Eranki, Chaitanya and Zhang, Christina and Park, Ji-Hwan and Hong, Raymond and Kalyani, Pranav and Kalyanaraman, Lochana and Gamare, Arsh and Bagad, Arnav and Esteva, Maria and others},
  journal={IEEE Transactions on Robotics},
  volume={40},
  pages={3322--3340},
  year={2024},
  publisher={IEEE}
}

@inproceedings{ettinger2021large,
  title={Large scale interactive motion forecasting for autonomous driving: The waymo open motion dataset},
  author={Ettinger, Scott and Cheng, Shuyang and Caine, Benjamin and Liu, Chenxi and Zhao, Hang and Pradhan, Sabeek and Chai, Yuning and Sapp, Ben and Qi, Charles R and Zhou, Yin and others},
  booktitle={Proceedings of the IEEE/CVF international conference on computer vision},
  pages={9710--9719},
  year={2021}
}

@article{hart1968formal,
  title={A formal basis for the heuristic determination of minimum cost paths},
  author={Hart, Peter E and Nilsson, Nils J and Raphael, Bertram},
  journal={IEEE transactions on Systems Science and Cybernetics},
  volume={4},
  number={2},
  pages={100--107},
  year={1968},
  publisher={IEEE}
}

@inproceedings{menda2019ensembledagger,
  title={Ensembledagger: A bayesian approach to safe imitation learning},
  author={Menda, Kunal and Driggs-Campbell, Katherine and Kochenderfer, Mykel J},
  booktitle={2019 IEEE/RSJ International Conference on Intelligent Robots and Systems (IROS)},
  pages={5041--5048},
  year={2019},
  organization={IEEE}
}

@inproceedings{wang2022feedback,
  title={Feedback-efficient active preference learning for socially aware robot navigation},
  author={Wang, Ruiqi and Wang, Weizheng and Min, Byung-Cheol},
  booktitle={2022 IEEE/RSJ international conference on intelligent robots and systems (IROS)},
  pages={11336--11343},
  year={2022},
  organization={IEEE}
}

@article{engesser2023autonomous,
  title={Autonomous delivery solutions for last-mile logistics operations: A literature review and research agenda},
  author={Engesser, Valeska and Rombaut, Evy and Vanhaverbeke, Lieselot and Lebeau, Philippe},
  journal={Sustainability},
  volume={15},
  number={3},
  pages={2774},
  year={2023},
  publisher={MDPI}
}

@article{liu2025service,
  title={Service robots in my workplace: Effects of employee-service robot co-work experiences on psychological empowerment},
  author={Liu, Xin and Zhang, Lu and Zhu, Tengteng},
  journal={Journal of Hospitality Marketing \& Management},
  volume={34},
  number={2},
  pages={175--203},
  year={2025},
  publisher={Taylor \& Francis}
}

@article{tuomi2021applications,
  title={Applications and implications of service robots in hospitality},
  author={Tuomi, Aarni and Tussyadiah, Iis P and Stienmetz, Jason},
  journal={Cornell Hospitality Quarterly},
  volume={62},
  number={2},
  pages={232--247},
  year={2021},
  publisher={SAGE Publications Sage CA: Los Angeles, CA}
}

@inproceedings{ross2011reduction,
  title={A reduction of imitation learning and structured prediction to no-regret online learning},
  author={Ross, St{\'e}phane and Gordon, Geoffrey and Bagnell, Drew},
  booktitle={Proceedings of the fourteenth international conference on artificial intelligence and statistics},
  pages={627--635},
  year={2011},
  organization={JMLR Workshop and Conference Proceedings}
}

@article{ziegler2019fine,
  title={Fine-tuning language models from human preferences},
  author={Ziegler, Daniel M and Stiennon, Nisan and Wu, Jeffrey and Brown, Tom B and Radford, Alec and Amodei, Dario and Christiano, Paul and Irving, Geoffrey},
  journal={arXiv preprint arXiv:1909.08593},
  year={2019}
}

@article{celemin2022interactive,
  title={Interactive imitation learning in robotics: A survey},
  author={Celemin, Carlos and P{\'e}rez-Dattari, Rodrigo and Chisari, Eugenio and Franzese, Giovanni and de Souza Rosa, Leandro and Prakash, Ravi and Ajanovi{\'c}, Zlatan and Ferraz, Marta and Valada, Abhinav and Kober, Jens and others},
  journal={Foundations and Trends{\textregistered} in Robotics},
  volume={10},
  number={1-2},
  pages={1--197},
  year={2022},
  publisher={Now Publishers, Inc.}
}

@article{wei2025ground,
  title={Ground slow, move fast: A dual-system foundation model for generalizable vision-and-language navigation},
  author={Wei, Meng and Wan, Chenyang and Peng, Jiaqi and Yu, Xiqian and Yang, Yuqiang and Feng, Delin and Cai, Wenzhe and Zhu, Chenming and Wang, Tai and Pang, Jiangmiao and others},
  journal={arXiv preprint arXiv:2512.08186},
  year={2025}
}

@inproceedings{
    rafailov2023direct,
    title={Direct Preference Optimization: Your Language Model is Secretly a Reward Model},
    author={Rafael Rafailov and Archit Sharma and Eric Mitchell and Christopher D Manning and Stefano Ermon and Chelsea Finn},
    booktitle={Thirty-seventh Conference on Neural Information Processing Systems},
    year={2023},
    url={https://arxiv.org/abs/2305.18290}
}

@article{chen2025socialnav,
  title={SocialNav: Training Human-Inspired Foundation Model for Socially-Aware Embodied Navigation},
  author={Chen, Ziyi and Guo, Yingnan and Chu, Zedong and Luo, Minghua and Shen, Yanfen and Sun, Mingchao and Hu, Junjun and Xie, Shichao and Yang, Kuan and Shi, Pei and others},
  journal={arXiv preprint arXiv:2511.21135},
  year={2025}
}

@inproceedings{pan2020zero,
  title={Zero-shot imitation learning from demonstrations for legged robot visual navigation},
  author={Pan, Xinlei and Zhang, Tingnan and Ichter, Brian and Faust, Aleksandra and Tan, Jie and Ha, Sehoon},
  booktitle={2020 IEEE International Conference on Robotics and Automation (ICRA)},
  pages={679--685},
  year={2020},
  organization={IEEE}
}

@article{argall2009survey,
  title={A survey of robot learning from demonstration},
  author={Argall, Brenna D and Chernova, Sonia and Veloso, Manuela and Browning, Brett},
  journal={Robotics and autonomous systems},
  volume={57},
  number={5},
  pages={469--483},
  year={2009},
  publisher={Elsevier}
}

@inproceedings{hu2023planning,
  title={Planning-oriented autonomous driving},
  author={Hu, Yihan and Yang, Jiazhi and Chen, Li and Li, Keyu and Sima, Chonghao and Zhu, Xizhou and Chai, Siqi and Du, Senyao and Lin, Tianwei and Wang, Wenhai and others},
  booktitle={Proceedings of the IEEE/CVF conference on computer vision and pattern recognition},
  pages={17853--17862},
  year={2023}
}

@article{arntz2023assessment,
  title={Assessment of readiness of a traffic environment for autonomous delivery robots},
  author={Arntz, EM and Van Duin, JHR and Van Binsbergen, AJ and Tavasszy, LA and Klein, T},
  journal={Frontiers in Future Transportation},
  volume={4},
  pages={1102302},
  year={2023},
  publisher={Frontiers Media SA}
}

@article{peng2023learning,
  title={Learning from active human involvement through proxy value propagation},
  author={Peng, Zhenghao Mark and Mo, Wenjie and Duan, Chenda and Li, Quanyi and Zhou, Bolei},
  journal={Advances in neural information processing systems},
  volume={36},
  pages={77969--77992},
  year={2023}
}

@inproceedings{cai2025predictive,
  title={Predictive Preference Learning from Human Interventions},
  author={Cai, Haoyuan and Peng, Zhenghao and Zhou, Bolei},
  booktitle={Advances in Neural Information Processing Systems},
  year={2025}
}

@inproceedings{shah2023gnm,
  title={Gnm: A general navigation model to drive any robot},
  author={Shah, Dhruv and Sridhar, Ajay and Bhorkar, Arjun and Hirose, Noriaki and Levine, Sergey},
  booktitle={2023 IEEE International Conference on Robotics and Automation (ICRA)},
  pages={7226--7233},
  year={2023},
  organization={IEEE}
}

@article{shah2023vint,
  title={ViNT: A foundation model for visual navigation},
  author={Shah, Dhruv and Sridhar, Ajay and Dashora, Nitish and Stachowicz, Kyle and Black, Kevin and Hirose, Noriaki and Levine, Sergey},
  journal={arXiv preprint arXiv:2306.14846},
  year={2023}
}

@inproceedings{sridhar2024nomad,
  title={Nomad: Goal masked diffusion policies for navigation and exploration},
  author={Sridhar, Ajay and Shah, Dhruv and Glossop, Catherine and Levine, Sergey},
  booktitle={2024 IEEE International Conference on Robotics and Automation (ICRA)},
  pages={63--70},
  year={2024},
  organization={IEEE}
}

@article{liu2024citywalker,
  title={CityWalker: Learning Embodied Urban Navigation from Web-Scale Videos},
  author={Liu, Xinhao and Li, Jintong and Jiang, Yicheng and Sujay, Niranjan and Yang, Zhicheng and Zhang, Juexiao and Abanes, John and Zhang, Jing and Feng, Chen},
  journal={arXiv preprint arXiv:2411.17820},
  year={2024}
}

@article{bar2024navigation,
  title={Navigation world models},
  author={Bar, Amir and Zhou, Gaoyue and Tran, Danny and Darrell, Trevor and LeCun, Yann},
  journal={arXiv preprint arXiv:2412.03572},
  year={2024}
}

@inproceedings{sadigh2017active,
  title={Active Preference-Based Learning of Reward Functions},
  author={Dorsa Sadigh and Anca D. Dragan and S. Shankar Sastry and Sanjit A. Seshia},
  booktitle={Robotics: Science and Systems},
  year={2017},
}

@article{ho2016generative,
  title={Generative adversarial imitation learning},
  author={Ho, Jonathan and Ermon, Stefano},
  journal={Advances in neural information processing systems},
  volume={29},
  year={2016}
}

@inproceedings{ziebart2008maximum,
  title={Maximum entropy inverse reinforcement learning.},
  author={Ziebart, Brian D and Maas, Andrew L and Bagnell, J Andrew and Dey, Anind K and others},
  booktitle={Aaai},
  volume={8},
  pages={1433--1438},
  year={2008},
  organization={Chicago, IL, USA}
}

@article{liu2022flow,
  title={Flow straight and fast: Learning to generate and transfer data with rectified flow},
  author={Liu, Xingchao and Gong, Chengyue and Liu, Qiang},
  journal={arXiv preprint arXiv:2209.03003},
  year={2022}
}

@inproceedings{vasu2023fastvit,
  title={Fastvit: A fast hybrid vision transformer using structural reparameterization},
  author={Vasu, Pavan Kumar Anasosalu and Gabriel, James and Zhu, Jeff and Tuzel, Oncel and Ranjan, Anurag},
  booktitle={Proceedings of the IEEE/CVF international conference on computer vision},
  pages={5785--5795},
  year={2023}
}

@article{liu2025improving,
  title={Improving video generation with human feedback},
  author={Liu, Jie and Liu, Gongye and Liang, Jiajun and Yuan, Ziyang and Liu, Xiaokun and Zheng, Mingwu and Wu, Xiele and Wang, Qiulin and Xia, Menghan and Wang, Xintao and others},
  journal={arXiv preprint arXiv:2501.13918},
  year={2025}
}

@article{liu2025flow,
  title={Flow-grpo: Training flow matching models via online rl},
  author={Liu, Jie and Liu, Gongye and Liang, Jiajun and Li, Yangguang and Liu, Jiaheng and Wang, Xintao and Wan, Pengfei and Zhang, Di and Ouyang, Wanli},
  journal={arXiv preprint arXiv:2505.05470},
  year={2025}
}

@article{he2025seeing,
  title={From seeing to experiencing: Scaling navigation foundation models with reinforcement learning},
  author={He, Honglin and Ma, Yukai and Wu, Wayne and Zhou, Bolei},
  journal={arXiv preprint arXiv:2507.22028},
  year={2025}
}

@article{kretzschmar2016socially,
  title={Socially compliant mobile robot navigation via inverse reinforcement learning},
  author={Kretzschmar, Henrik and Spies, Markus and Sprunk, Christoph and Burgard, Wolfram},
  journal={The International Journal of Robotics Research},
  volume={35},
  number={11},
  pages={1289--1307},
  year={2016},
  publisher={SAGE Publications Sage UK: London, England}
}

@inproceedings{choi2020fast,
  title={Fast adaptation of deep reinforcement learning-based navigation skills to human preference},
  author={Choi, Jinyoung and Dance, Christopher and Kim, Jung-eun and Park, Kyung-sik and Han, Jaehun and Seo, Joonho and Kim, Minsu},
  booktitle={2020 IEEE International Conference on Robotics and Automation (ICRA)},
  pages={3363--3370},
  year={2020},
  organization={IEEE}
}

@article{knox2022models,
  title={Models of human preference for learning reward functions},
  author={Knox, W Bradley and Hatgis-Kessell, Stephane and Booth, Serena and Niekum, Scott and Stone, Peter and Allievi, Alessandro},
  journal={arXiv preprint arXiv:2206.02231},
  year={2022}
}

@article{thrun2002probabilistic,
  title={Probabilistic robotics},
  author={Thrun, Sebastian},
  journal={Communications of the ACM},
  volume={45},
  number={3},
  pages={52--57},
  year={2002},
  publisher={ACM New York, NY, USA}
}

@inproceedings{zhang2025creste,
      title={CREStE: Scalable Mapless Navigation with Internet Scale Priors and Counterfactual Guidance}, 
      author={Arthur Zhang and Harshit Sikchi and Amy Zhang and Joydeep Biswas},
      booktitle={Robotics: Science and Systems (RSS)},
      year={2025}
}

@article{cheng2024navila,
  title={Navila: Legged robot vision-language-action model for navigation},
  author={Cheng, An-Chieh and Ji, Yandong and Yang, Zhaojing and Gongye, Zaitian and Zou, Xueyan and Kautz, Jan and B{\i}y{\i}k, Erdem and Yin, Hongxu and Liu, Sifei and Wang, Xiaolong},
  journal={arXiv preprint arXiv:2412.04453},
  year={2024}
}

@article{dauner2024navsim,
  title={Navsim: Data-driven non-reactive autonomous vehicle simulation and benchmarking},
  author={Dauner, Daniel and Hallgarten, Marcel and Li, Tianyu and Weng, Xinshuo and Huang, Zhiyu and Yang, Zetong and Li, Hongyang and Gilitschenski, Igor and Ivanovic, Boris and Pavone, Marco and others},
  journal={Advances in Neural Information Processing Systems},
  volume={37},
  pages={28706--28719},
  year={2024}
}

@misc{nvidia2025physicalaiav,
title = {PhysicalAI Autonomous Vehicles Dataset},
author = {{NVIDIA Corporation}},
year = {2025},
howpublished = {\url{https://huggingface.co/datasets/nvidia/PhysicalAI-Autonomous-Vehicles}},
note = {Hugging Face dataset. Accessed: 2026-06-10}
}

@article{hu2022lora,
  title={Lora: Low-rank adaptation of large language models.},
  author={Hu, Edward J and Shen, Yelong and Wallis, Phillip and Allen-Zhu, Zeyuan and Li, Yuanzhi and Wang, Shean and Wang, Lu and Chen, Weizhu and others},
  journal={ICLR},
  volume={1},
  number={2},
  pages={3},
  year={2022}
}

@inproceedings{liao2025diffusiondrive,
  title={Diffusiondrive: Truncated diffusion model for end-to-end autonomous driving},
  author={Liao, Bencheng and Chen, Shaoyu and Yin, Haoran and Jiang, Bo and Wang, Cheng and Yan, Sixu and Zhang, Xinbang and Li, Xiangyu and Zhang, Ying and Zhang, Qian and others},
  booktitle={Proceedings of the Computer Vision and Pattern Recognition Conference},
  pages={12037--12047},
  year={2025}
}

@inproceedings{cai2025robot,
  title={Robot-Gated Interactive Imitation Learning with Adaptive Intervention Mechanism},
  author={Cai, Haoyuan and Peng, Zhenghao and Zhou, Bolei},
  booktitle={International Conference on Machine Learning},
  year={2025}
}

@article{bjorck2025gr00t,
  title={GR00T N1: An Open Foundation Model for Generalist Humanoid Robots},
  author={{NVIDIA} and Johan Bjorck and Fernando Castañeda and Nikita Cherniadev and Xingye Da and Runyu Ding and Linxi "Jim" Fan and Yu Fang and Dieter Fox and Fengyuan Hu and Spencer Huang and Joel Jang and Zhenyu Jiang and Jan Kautz and Kaushil Kundalia and Lawrence Lao and Zhiqi Li and Zongyu Lin and Kevin Lin and Guilin Liu and Edith Llontop and Loic Magne and Ajay Mandlekar and Avnish Narayan and Soroush Nasiriany and Scott Reed and You Liang Tan and Guanzhi Wang and Zu Wang and Jing Wang and Qi Wang and Jiannan Xiang and Yuqi Xie and Yinzhen Xu and Zhenjia Xu and Seonghyeon Ye and Zhiding Yu and Ao Zhang and Hao Zhang and Yizhou Zhao and Ruijie Zheng and Yuke Zhu},
  journal={arXiv preprint arXiv:2503.14734},
  year={2025}
}

@inproceedings{liu2025citywalker,
  title={Citywalker: Learning embodied urban navigation from web-scale videos},
  author={Liu, Xinhao and Li, Jintong and Jiang, Yicheng and Sujay, Niranjan and Yang, Zhicheng and Zhang, Juexiao and Abanes, John and Zhang, Jing and Feng, Chen},
  booktitle={Proceedings of the Computer Vision and Pattern Recognition Conference},
  pages={6875--6885},
  year={2025}
}

@article{chi2025diffusion,
  title={Diffusion policy: Visuomotor policy learning via action diffusion},
  author={Chi, Cheng and Xu, Zhenjia and Feng, Siyuan and Cousineau, Eric and Du, Yilun and Burchfiel, Benjamin and Tedrake, Russ and Song, Shuran},
  journal={The International Journal of Robotics Research},
  volume={44},
  number={10-11},
  pages={1684--1704},
  year={2025},
  publisher={Sage Publications Sage UK: London, England}
}

@article{shah2022gnm,
  title={Gnm: A general navigation model to drive any robot},
  author={Shah, Dhruv and Sridhar, Ajay and Bhorkar, Arjun and Hirose, Noriaki and Levine, Sergey},
  journal={arXiv preprint:2210.03370},
  year={2022}
}

@inproceedings{van2011reciprocal,
  title={Reciprocal n-body collision avoidance},
  author={Van Den Berg, Jur and Guy, Stephen J and Lin, Ming and Manocha, Dinesh},
  booktitle={Robotics research: the 14th international symposium ISRR},
  pages={3--19},
  year={2011},
  organization={Springer}
}

@inproceedings{wu2025towards,
  title={Towards autonomous micromobility through scalable urban simulation},
  author={Wu, Wayne and He, Honglin and Zhang, Chaoyuan and He, Jack and Zhao, Seth Z and Gong, Ran and Li, Quanyi and Zhou, Bolei},
  booktitle={Proceedings of the Computer Vision and Pattern Recognition Conference},
  pages={27553--27563},
  year={2025}
}

@article{ouyang2022training,
  title={Training language models to follow instructions with human feedback},
  author={Ouyang, Long and Wu, Jeffrey and Jiang, Xu and Almeida, Diogo and Wainwright, Carroll L. and Mishkin, Pamela and Zhang, Chong and Agarwal, Sandhini and Slama, Katarina and Ray, Alex and others},
  journal={arXiv preprint arXiv:2203.02155},
  year={2022}
}

@inproceedings{deng2009imagenet,
  title={Imagenet: A large-scale hierarchical image database},
  author={Deng, Jia and Dong, Wei and Socher, Richard and Li, Li-Jia and Li, Kai and Fei-Fei, Li},
  booktitle={2009 IEEE conference on computer vision and pattern recognition},
  pages={248--255},
  year={2009},
  organization={Ieee}
}

@article{cai2025navdp,
  title={Navdp: Learning sim-to-real navigation diffusion policy with privileged information guidance},
  author={Cai, Wenzhe and Peng, Jiaqi and Yang, Yuqiang and Zhang, Yujian and Wei, Meng and Wang, Hanqing and Chen, Yilun and Wang, Tai and Pang, Jiangmiao},
  journal={arXiv preprint arXiv:2505.08712},
  year={2025}
}

@article{he2026learning,
  title={Learning Sidewalk Autopilot from Multi-Scale Imitation with Corrective Behavior Expansion},
  author={He, Honglin and Ma, Yukai and Squicciarini, Brad and Wu, Wayne and Zhou, Bolei},
  journal={arXiv preprint arXiv:2603.22527},
  year={2026}
}
\clearpage

\appendix
\begin{flushleft}
{\Large\textbf{Appendix}}
\end{flushleft}

We organize the appendix as follows. We first introduce the demonstration video in Sec.~\ref{sec:video}, followed by additional real-world results in Sec.~\ref{sec:real}. We then provide additional simulation results in Sec.~\ref{sec:sim}, open-loop evaluation results in Sec.~\ref{sec:openloop}, and the implementation details in Sec.~\ref{sec:app-impl}. We also provide an ethics statement in Sec.~\ref{sec:ethics}, where we describe the safety protocol used during real-world deployment.

\section{Demonstration Video}
\label{sec:video}

We highly recommend watching our supplementary video for detailed demonstrations. The video presents a broad set of real-world experiments that thoroughly evaluate the performance of \ModelName\ across diverse sidewalk environments. It consists of four sections:

\paragraph{1. \ModelName\ capabilities demonstration} \ModelName\ demonstrates robust navigation in complex real-world sidewalk environments. It successfully negotiates narrow passages, cluttered layouts, and broken curbs while maintaining safe and socially compliant behaviors, including effective obstacle avoidance and pedestrian awareness.

\paragraph{2. Long-horizon sidewalk navigation} \ModelName\ completes GPS-guided long-horizon navigation with only a few human interventions. It maintains stable sidewalk lane keeping and consistent goal progress over time, while remaining robust to lighting changes and transient disturbances in challenging sidewalk environments.

\paragraph{3. Comparison with SOTA methods} We present side-by-side real-world evaluations against representative state-of-the-art methods under the same setting. The comparisons highlight the advantages of \ModelName\ in trajectory smoothness, navigation stability, and safety.

\paragraph{4. Cross-embodiment generality} \ModelName\ transfers effectively across different robot platforms both without finetuning and with only a few embodiment-specific examples. It preserves reasonable navigation behaviors under changes in platform dynamics and sensing configurations, demonstrating strong generalization and rapid adaptation across embodiments.

\section{Real-World Closed-Loop Results}
\label{sec:real}

\subsection{Robot hardware setup}
\label{sec:app-robot}

\begin{wrapfigure}{r}{0.5\textwidth}
\centering
\vspace{-4mm}
\includegraphics[width=0.48\textwidth]{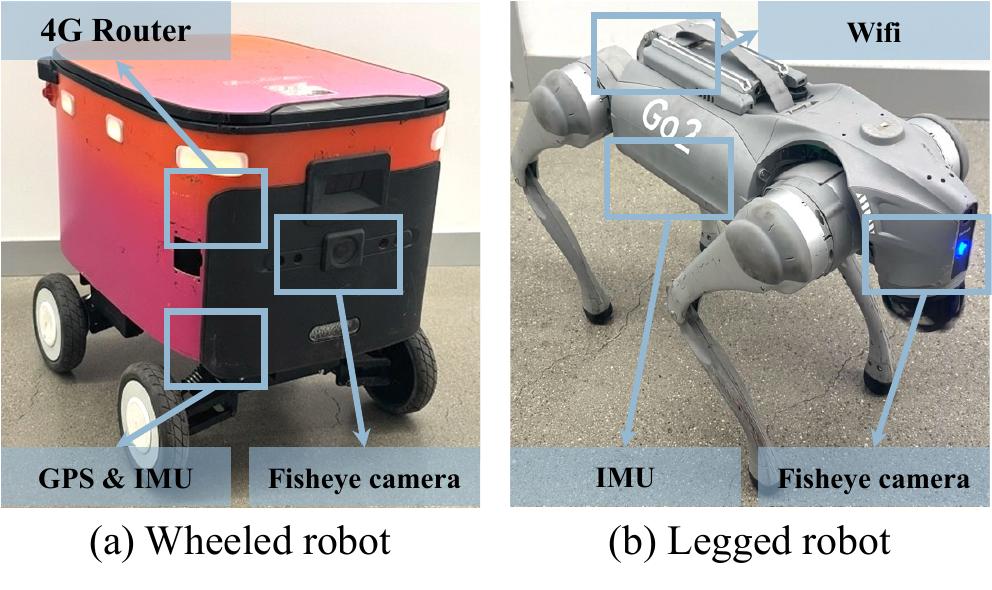}
\vspace{-3mm}
\caption{\textbf{Robot platforms used in real-world experiments.} We use a wheeled robot for the majority of experiments, and deploy a legged robot to assess cross-embodiment generalization.}
\label{fig:robotsetup}
\vspace{-3mm}
\end{wrapfigure}

As illustrated in Fig.~\ref{fig:robotsetup}, we evaluate \ModelName\ on two robot platforms for real-world experiments. We use a wheeled robot for the majority of real-world experiments, as it represents the primary deployment platform in our sidewalk navigation setting. In addition, we deploy \ModelName\ on a legged robot to assess cross-embodiment generalization, as discussed in Sec.~\ref{sec:app-cross}. This setup allows us to evaluate whether the learned navigation behaviors remain effective when transferred to platforms with different dynamics, actuation constraints, and sensing configurations.
For both platforms, we use onboard sensors to capture visual observations during navigation. The sensor streams are transmitted over a network to an external compute unit, where the policy performs inference and predicts navigation actions. These actions are then sent back to the robot for execution at 10-20~FPS. This deployment pipeline enables us to use the same policy interface across different platforms while keeping the onboard robot setup lightweight.
During deployment, we use a laptop as the operator interface to visualize the live camera stream, predicted trajectory, robot state, and system status. A joystick controller is connected to the deployment laptop and is used by the safety operator to take over control whenever intervention is required. Each intervention is automatically recorded by monitoring the control-status topic, which indicates whether the robot is under autonomous policy control or manual joystick control. The corresponding timestamps are synchronized with the logged sensor streams and robot states, allowing us to label intervention segments and compute intervention-based evaluation metrics.

\begin{wraptable}{r}{0.50\textwidth} 
    \vspace{-3.5mm}
    \centering
    {\footnotesize
    \setlength{\tabcolsep}{3.5pt}
    \resizebox{\linewidth}{!}{
    \begin{tabular}{l cccc}
        \toprule
        Method 
        & IR~$\downarrow$ 
        & NIR~$\downarrow$ 
        & DCR~$\uparrow$ 
        & TCR~$\uparrow$ \\
        \midrule
        \ModelNameIL
        & 0.055 & 0.106 & 0.922 & 0.937 \\
        \ModelNameFull
        & \textbf{0.017} 
        & \textbf{0.049} 
        & \textbf{0.949} 
        & \textbf{0.951} \\
        \bottomrule
    \end{tabular}}}
    \caption{\textbf{Additional closed-loop evaluation in real-world sidewalk environments.}}
    \vspace{-4mm}
    \label{sup--tab:real-world-closed-loop}
\end{wraptable}

We conduct an additional real-world closed-loop evaluation across more diverse scenarios to further validate the robustness of our policy. We deploy the robot in 5 urban blocks during dusk and nighttime, resulting in 10 total trials and covering approximately 17.8~km of real-world sidewalk navigation. Due to the unstable deployment behavior of several baseline methods in long-horizon real-world settings, we do not repeatedly deploy them in this expanded evaluation for safety reasons. As given in Tab.~\ref{sup--tab:real-world-closed-loop}, the results show that \ModelNameFull\ reduces both IR and NIR while improving DCR and TCR, demonstrating better real-world robustness and social compliance.

\begin{figure}[t!]
\centering
\includegraphics[width=\textwidth]{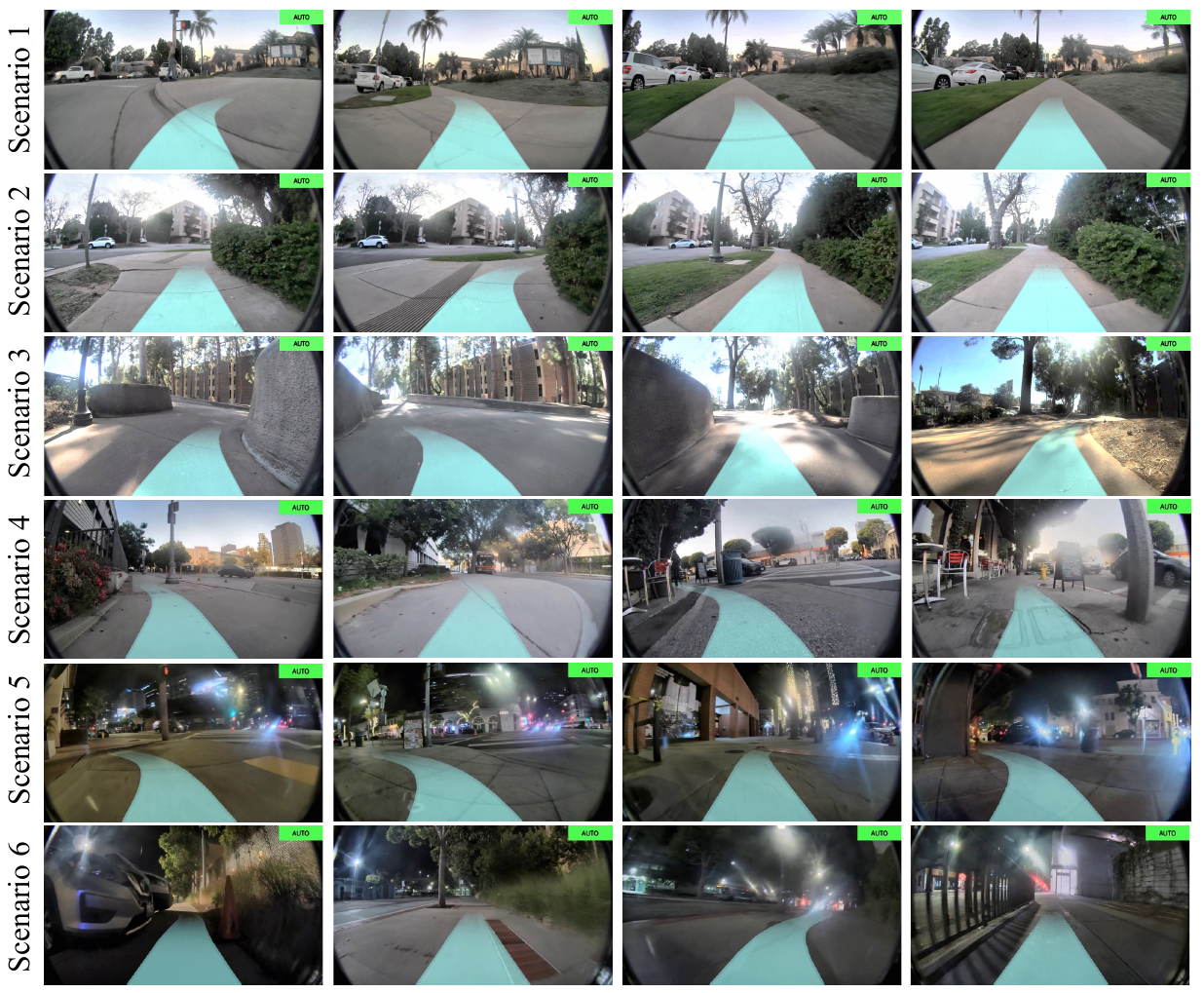}
\caption{\textbf{Sidewalk lane keeping in diverse real-world scenes.}
\ModelName\ maintains a stable position within the sidewalk under varying appearances and layouts, producing smooth, safe trajectories.}
\label{fig:lanekeeping}
\vspace{-3mm}
\end{figure}

\subsection{Long-horizon sidewalk navigation}
\label{sec:longhorizon}

As illustrated in Fig.~\ref{fig:long}, Fig.~\ref{fig:long-2}, as well as Sec.~2 of the supplementary video, \ModelName\ demonstrates robust long-horizon sidewalk navigation across diverse real-world conditions. Fig.~\ref{fig:long} presents a kilometer-scale daytime route, where the robot maintains consistent goal progress over an extended distance while performing stable sidewalk lane keeping, smooth turning, and obstacle avoidance. Fig.~\ref{fig:long-2} further evaluates \ModelName\ in a nighttime route with low-light regions and complex illumination changes, showing that the policy remains robust under degraded visual conditions. Across both cases, \ModelName\ generates stable and socially compliant behaviors, avoids obstacles, and recovers from minor deviations without human intervention. Sec.~2 of the supplementary video provides the full video demonstrations of these two long-horizon cases, and we highly recommend readers to watch the video for a more detailed visualization of the navigation behaviors and qualitative results.

\subsection{Sidewalk lane keeping}
\label{sec:lanekeeping}

As illustrated in Fig.~\ref{fig:lanekeeping}, \ModelName\ reliably tracks the sidewalk corridor across a wide range of layouts, including narrow passages, curb-constrained segments, texture changes, etc. The policy keeps a stable lateral offset from boundaries (\eg, curbs, grass edges, sidewalk boundaries), resulting in smooth behaviors. When the visual appearance or sidewalk width changes, the robot quickly re-centers and continues forward progress without requiring teleoperation or explicit re-initialization.

\subsection{Obstacle avoidance}
\label{sec:obstacle}

\begin{figure}[t!]
\centering
\includegraphics[width=\textwidth]{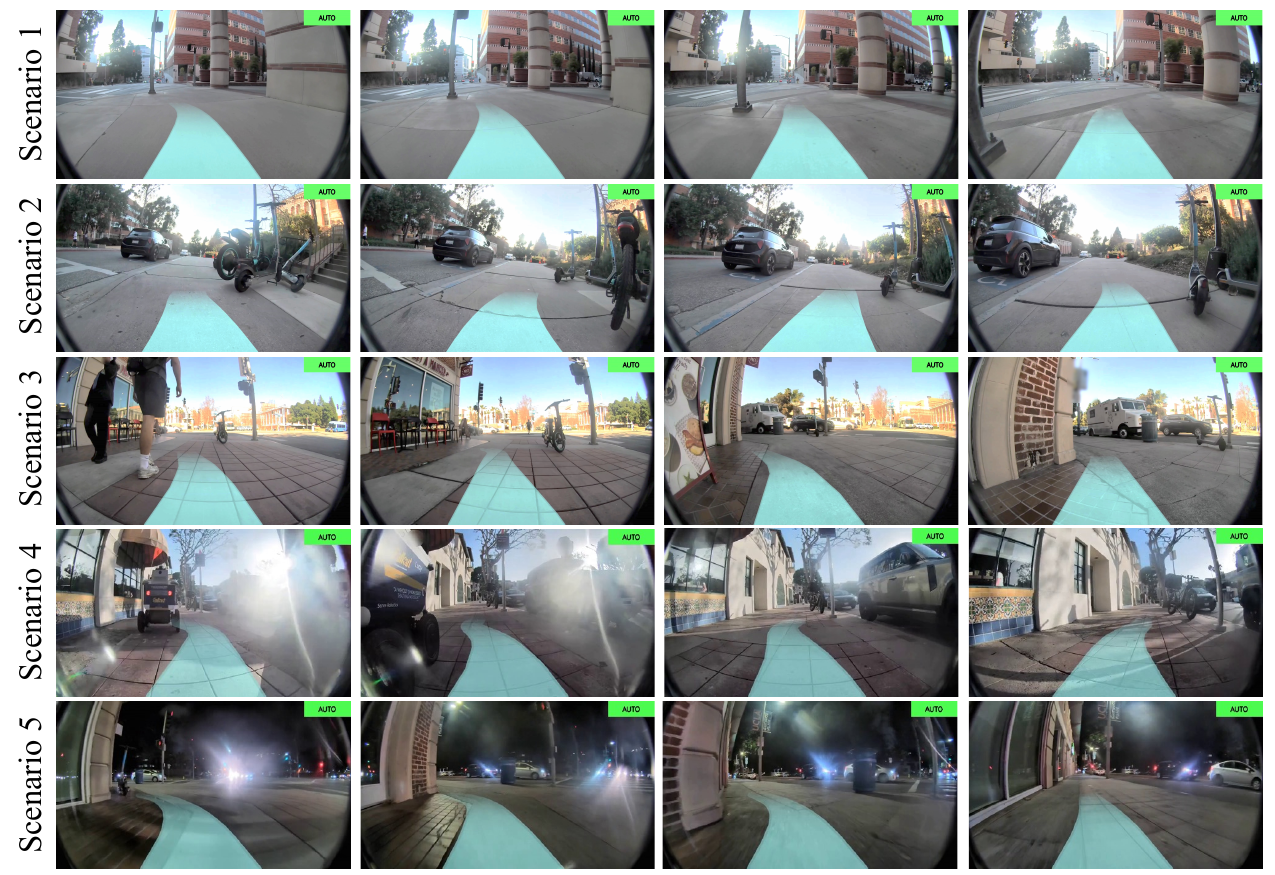}
\caption{\textbf{Reactive obstacle avoidance in cluttered sidewalks.} \ModelName\ selects safe bypass maneuvers around static obstacles and returns to the nominal path after passing them.}
\label{fig:obstacle}
\vspace{-3mm}
\end{figure}

As illustrated in Fig.~\ref{fig:obstacle}, \ModelName\ performs robust obstacle avoidance in cluttered sidewalk environments with common static obstacles (\eg, trash bins, cones, parked micromobility devices, etc.). The robot proactively slows down when clearance becomes tight, chooses a conservative side to pass, and maintains a collision-free margin while preserving forward progress. Once the obstacle is bypassed, the policy executes a smooth transition back to the sidewalk centerlines, rather than oscillating near the boundaries.

\subsection{Pedestrian awareness}
\label{sec:pedestrian}

\begin{figure}[htbp]
\centering
\includegraphics[width=\textwidth]{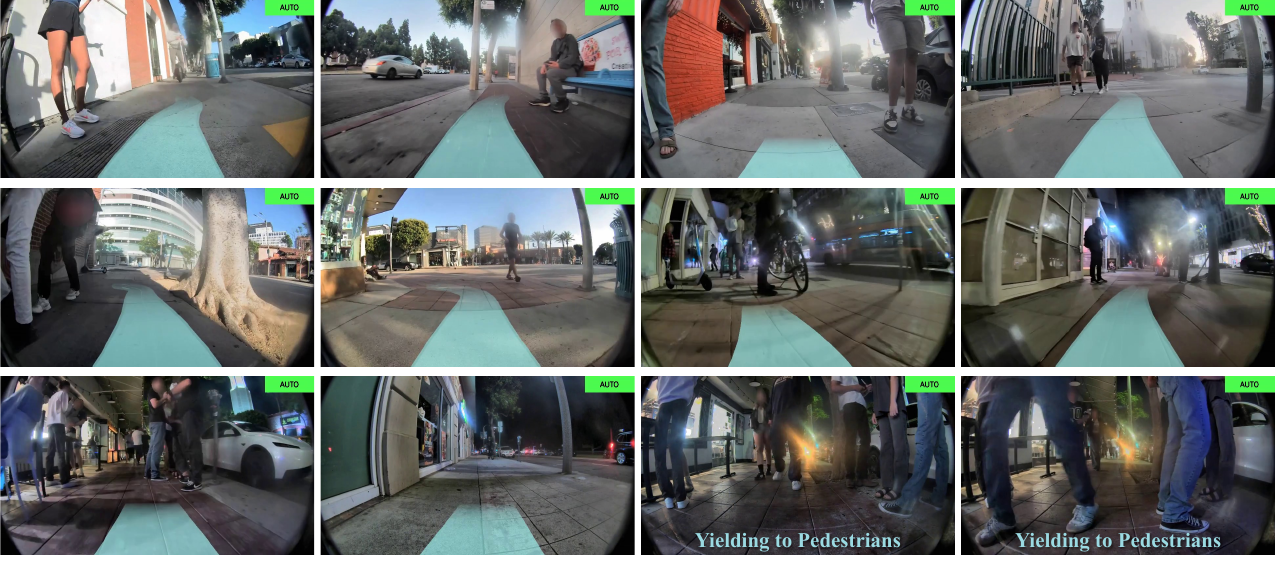}
\caption{\textbf{Pedestrian-aware navigation without interventions.}
\ModelName\ keeps comfortable clearance, yields in crowded situations, and re-centers once the path is clear.}
\label{fig:pedestrian}
\vspace{-3mm}
\end{figure}

As illustrated in Fig.~\ref{fig:pedestrian}, \ModelName\ exhibits pedestrian-aware behaviors that are consistent with socially compliant navigation. In crowded scenes, the robot reduces speed, maintains a comfortable clearance, and yields when the available space is insufficient for a safe pass. When pedestrians move away or the corridor opens up, the policy accelerates smoothly and re-centers, avoiding abrupt turns or any aggressive behaviors.

\subsection{Cross-embodiment generalization}
\label{sec:app-cross}

Since our policy outputs \emph{waypoints} (rather than platform-specific low-level controls), it can be deployed across robot embodiments with minimal interface changes. Concretely, the same \ModelName\ predicts short-horizon waypoints in a robot-centric frame, which are then tracked by a platform-specific controller (\eg, a PD controller for the wheeled base and a velocity tracker for the legged robot). This modeling enables transferring the learned navigation behavior to other robot platforms without additional fine-tuning or few-shot demonstrations. Specifically, we collect 30 minutes of preference data on the Go2 platform and finetune the policy with the same preference-learning objective. More quantitative results are provided in Sec.~\ref{supp:preference-open}. As illustrated in the final section of the demonstration video, we provide two deployment examples on the Unitree Go2 legged robot, showing that \ModelName\ can preserve stable sidewalk-following and obstacle-avoidance behaviors across different robot embodiments. 

\subsection{Typical failure cases}
\label{sec:failure}

\begin{wrapfigure}{r}{0.5\textwidth}
\centering
\vspace{-4mm}
\includegraphics[width=\linewidth]{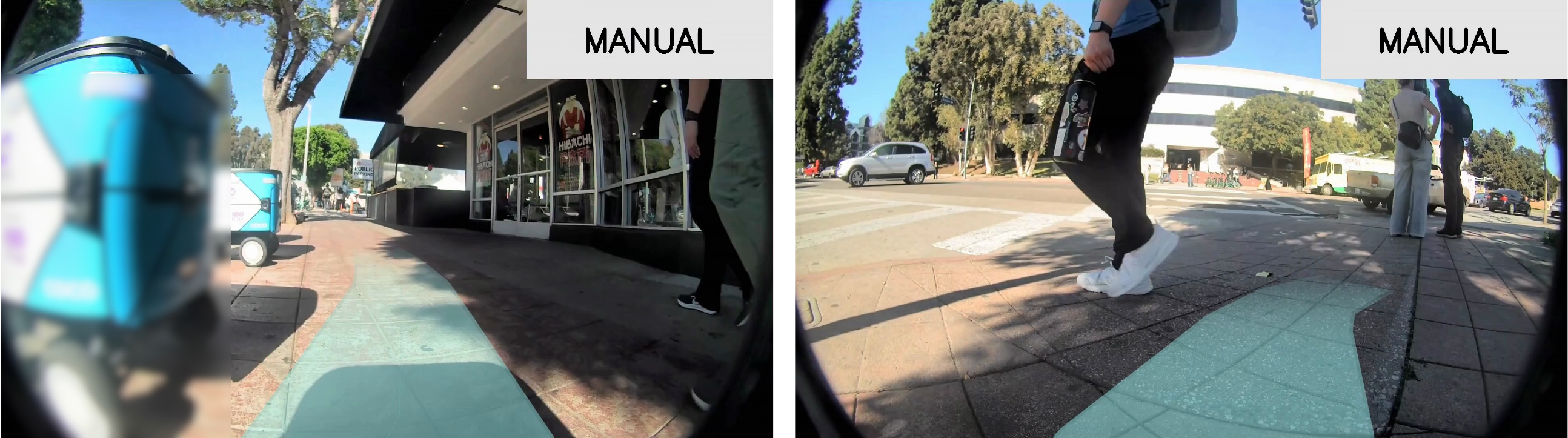}
\caption{\textbf{Representative failure cases in crowded or constrained scenes.}
When faced with ambiguous right-of-way or insufficient clearance, \ModelName\ may be overly conservative and stop, requiring human intervention to continue.}
\label{fig:failure}
\end{wrapfigure}

As shown in Fig.~\ref{fig:failure}, while \ModelName\ generally exhibits pedestrian-aware and socially compliant behaviors, we also observe suboptimal behaviors in challenging edge cases. \textbf{(1) Narrow passage with pedestrians:} in a narrow space, the policy fails to generate a feasible passing behavior between the other robots and nearby pedestrians and instead becomes stuck in an overly conservative state, repeatedly stopping until an intervention changes the scene or the robot state. \textbf{(2) Crosswalk / traffic-light scenario with a pedestrian present:} when a pedestrian remains within the interaction zone, the policy defaults to stopping and does not execute an alternative safe behavior (\eg, edging forward to improve visibility or re-positioning), leading to deadlock. These cases highlight the need for stronger intent reasoning and explicit deadlock resolution strategies in dense human-robot interaction.

\section{Simulation Closed-Loop Results}
\label{sec:sim}

\begin{figure}[htbp]
\centering
\includegraphics[width=\textwidth]{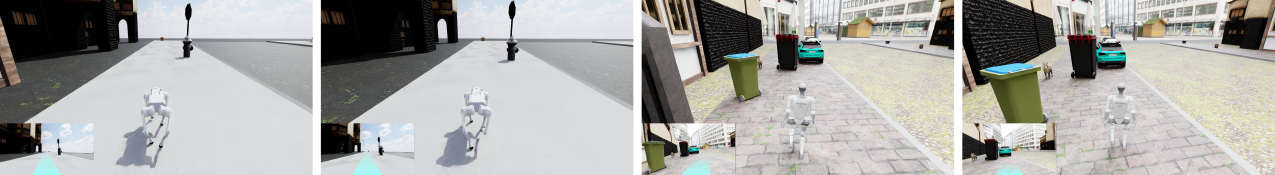}
\caption{\textbf{Cross-embodiment generalization in simulated environments.} }
\label{fig:Cross-Sim}
\vspace{-3mm}
\end{figure}

We further demonstrate the cross-embodiment generalization ability of \ModelName\ in closed-loop simulation, as demonstrated in Fig.~\ref{fig:Cross-Sim}. Since \ModelName\ predicts platform-agnostic waypoints in the robot-centric frame, the same high-level policy can be deployed on different robot configurations by changing only the low-level tracking controller. We deploy the policy on two legged embodiments, Unitree Go2 and Unitree G1, to demonstrate cross-embodiment generalization in closed-loop simulation. Since \ModelName\ outputs short-horizon waypoints in the robot-centric frame, we use an embodiment-specific tracking module to convert the predicted waypoints into low-level locomotion commands. For each robot, we train a separate locomotion policy in IsaacLab~\cite{mittal2025isaac} to track desired base velocities and local waypoint targets. This setup allows us to test whether the learned navigation behavior can transfer across substantially different robot dynamics.

\section{Open-Loop Results}
\label{sec:openloop}

In this section, we provide additional open-loop evaluation results to complement the results in the main paper.
Open-loop evaluation measures whether the predicted future trajectories match human teleoperation trajectories under the same visual observations. We analyze the effectiveness of each key component in the following sections.




\subsection{Effectiveness of action representation}
\label{supp:action-rep}

Tab.~\ref{tab:open-loop-action-rep} summarizes an ablation study on different trajectory parameterizations for open-loop prediction on SideWalks-300. 

\begin{wraptable}{r}{0.48\textwidth}
\vspace{-3.5mm}
\centering
\footnotesize
\setlength{\tabcolsep}{3pt}
\renewcommand{\arraystretch}{1.05}
\resizebox{\linewidth}{!}{%
\begin{tabular}{lcccc}
\toprule
\textbf{Method} & minMOE~$\downarrow$ & minADE~$\downarrow$ & L2~$\downarrow$ & mAP~$\uparrow$ \\
\midrule
Discrete &7.68 &0.58 &1.59 &0.82 \\ 
Regression & 6.77 & \textbf{0.46} & 1.73 & 0.81 \\ 
Diffusion & 6.96 & 0.63 & 1.71 & 0.77 \\ 
\midrule 
\ours \ModelNameIL\ & \textbf{6.63} & 0.49 & \textbf{1.04} & \textbf{0.87} \\
\bottomrule
\end{tabular}}%
\caption{\textbf{Action representation comparison on open-loop benchmark.}}
\label{tab:open-loop-action-rep}
\vspace{-5mm}
\end{wraptable}

\textbf{Discrete (K-means~+~Scoring):} we first cluster expert trajectories into $K$ anchors and train a scoring network to select one anchor at inference time. While this provides a multi-modal hypothesis set, the selected prototype can be overly coarse, yielding suboptimal tracking quality when converted into controls.

\textbf{Regression (S2E~\cite{he2025seeing}):} direct regression predicts a single trajectory, which tends to average over multiple plausible futures and can underperform in ambiguous scenes.

\textbf{Diffusion (DiffusionDrive~\cite{liao2025diffusiondrive}):} diffusion improves multi-modality but may introduce high-frequency noise, leading to less stable predicted plans. 
\textbf{Ours (anchored flow matching):} by combining anchor-based multi-modality with continuous refinement, our method achieves the best overall performance across metrics.

\subsection{Effectiveness of preference learning}
\label{supp:preference-open}

\begin{table*}[htbp] 
\centering
\footnotesize
\setlength{\tabcolsep}{3pt}
\renewcommand{\arraystretch}{1.05}

\begin{minipage}{0.48\linewidth}
\centering
\resizebox{\linewidth}{!}{%
\begin{tabular}{lcccc}
\toprule
\textbf{Method} & minMOE~$\downarrow$ & minADE~$\downarrow$ & L2~$\downarrow$ & mAP~$\uparrow$ \\
\midrule
GNM            &7.31 &0.63 &1.32  &-     \\
        ViNT           &8.31 &2.61 &4.47  &-    \\
        NoMaD         &11.10 &1.72 &3.17  &-     \\
        CityWalker     &10.83 &2.04 &3.59 &0.37\\

        CityWalker-HP     &9.71   &1.55   &2.93   &0.46    \\

        \ours \ModelNameIL\ &4.63 &1.17 &3.11 &0.53  \\

        \ours \ModelNameFull\ & \textbf{3.64} &\textbf{0.61}  &\textbf{1.67} &\textbf{0.79}  \\
        
        \bottomrule
\end{tabular}}%
\caption{\textbf{Open-loop benchmark results on robot-specific dataset (Wheeled robot).}}
\label{tab:open-loop-pref-coco-1}
\end{minipage}\hfill
\begin{minipage}{0.48\linewidth}
\centering
\resizebox{\linewidth}{!}{%
\begin{tabular}{lcccc}
\toprule
\textbf{Method} & minMOE~$\downarrow$ & minADE~$\downarrow$ & L2~$\downarrow$ & mAP~$\uparrow$ \\
\midrule
GNM                    & 31.99 & 2.13 & 3.79 & -    \\
        ViNT                   & 11.68 & 1.34 & 2.48 & -   \\
        NoMaD                  & 22.81 & 1.34 & 2.59 & -   \\
        CityWalker             & 12.49 & 1.01 & 1.99 & 0.19  \\
        CityWalker-HP          & 17.17  & 0.75 & 1.70  & 0.23    \\
        \ours \ModelNameIL\    & 3.59 & 0.53 & 2.99 & 0.27 \\
        \ours FlowPilot-HF & \textbf{3.24} & \textbf{0.21} & \textbf{0.91} & \textbf{0.74} \\
\bottomrule
\end{tabular}}%
\caption{\textbf{Open-loop benchmark results on robot-specific dataset (Legged robot).}}
\label{tab:open-loop-pref-go2-2}
\end{minipage}
\end{table*}

Tab.~\ref{tab:open-loop-pref-coco-1} and Tab.~\ref{tab:open-loop-pref-go2-2} quantify the impact of preference learning on robot-specific open-loop benchmarks for the wheeled and legged platforms, respectively. Specifically, we collect approximately 2 hours of data for each robot and split it into two splits for training and testing. We first pretrain \ModelName\ on the large-scale dataset and then fine-tune it on the robot-specific training split; preference learning is performed using human intervention pairs collected on the corresponding platform. As shown in the tables, preference fine-tuning consistently improves trajectory quality and action-level alignment (lower minMOE~/~minADE and L2, higher mAP), indicating better adaptation to safety-critical interactions and ambiguous situations in real deployments. We also fine-tune CityWalker~\cite{liu2024citywalker} to get CityWalker-HP on the same robot-specific training split. Notably, \ModelName\ achieves substantially stronger performance, suggesting that our model provides richer priors and more transferable navigation knowledge, which preference learning further refines to better match human preferences.

\section{Implementation Details}
\label{sec:app-impl}

\subsection{Pretraining dataset}
\label{sec:pretrain}

\begin{figure}[htbp]
\centering
\includegraphics[width=\columnwidth]{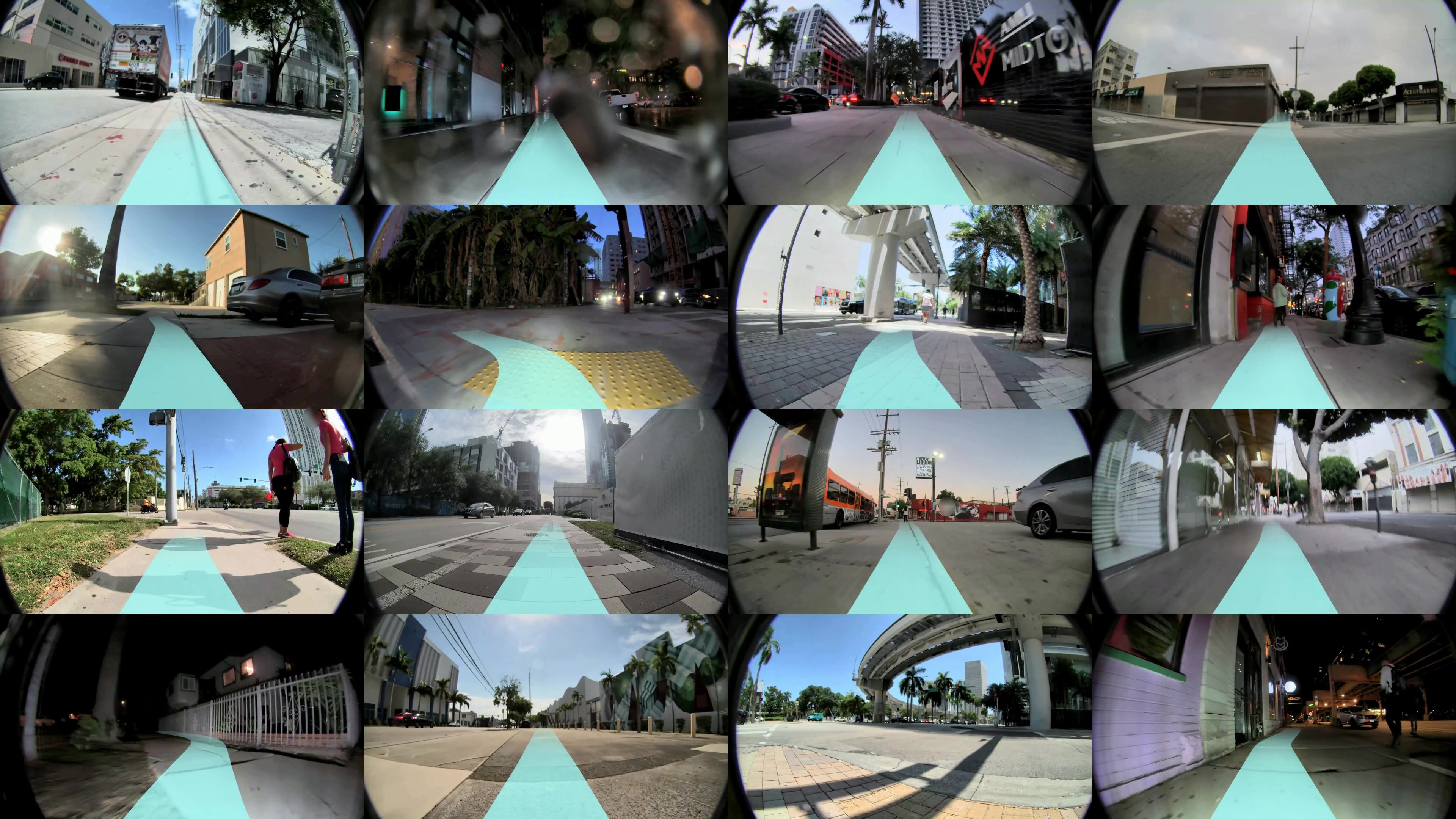}
\caption{\textbf{Overview of the large-scale pretraining dataset.}
We pretrain \ModelName\ on $\sim$300 hours of diverse real-world navigation data spanning varied lighting, weather, and scene layouts, including frequent interactions with obstacles and pedestrians.}
\label{fig:pretrain}
\end{figure}

As illustrated in Fig.~\ref{fig:pretrain}, our pretraining dataset contains approximately 300 hours of real-world robot navigation collected across diverse environments and conditions. It covers substantial variation in visual appearance (\eg, time of day, strong sunlight/glare, shadows, and different camera exposures), sidewalk layouts (\eg, curved segments, ramps, curbs, and intersections), and dynamic complexity. The data includes frequent interactions with static obstacles and moving agents such as pedestrians, micromobility robots, and animals, providing rich information for learning.

\subsection{Expert intervention data collection in real-world sidewalk environments}
\label{sec:app-human}

\begin{figure}[t!]
\centering
\includegraphics[width=\textwidth]{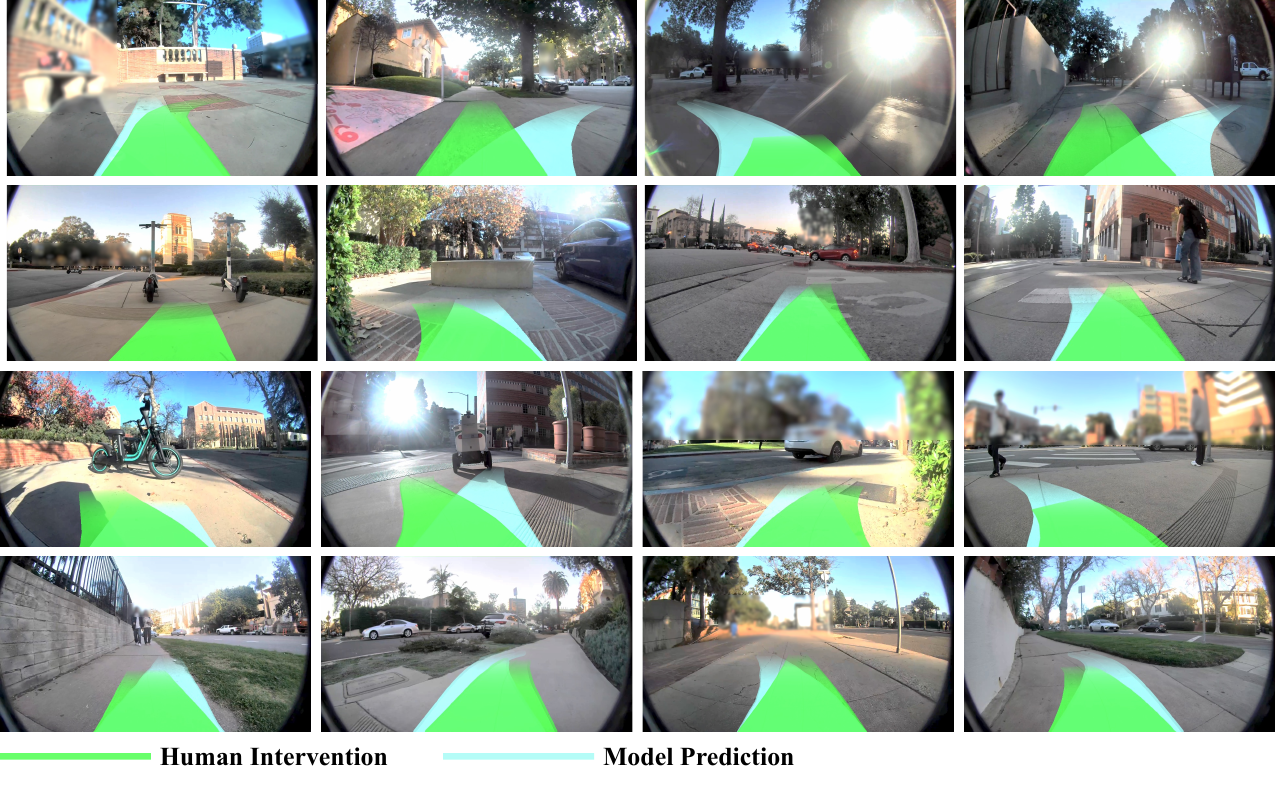}
\caption{\textbf{Qualitative comparison between \ModelName\ predictions and human interventions.}
We visualize representative real-world frames with the model-predicted path (cyan) and the trajectory executed during human intervention (green). Across diverse sidewalk scenes, the intervention traces highlight where humans deviate from the model to ensure safety and progress.}
\label{fig:intervention}
\end{figure}

After pretraining, we deploy \ModelName\ in real-world navigation with a human-in-the-loop safety protocol to collect intervention data. At each timestep, the policy predicts a short-horizon trajectory, which is visualized and treated as the \emph{candidate} control plan. A human operator monitors the robot and intervenes whenever the predicted plan is unsafe or likely to fail (\eg, insufficient clearance, imminent collision, or deadlock). During an intervention, we record a paired sample consisting of the model prediction and the human-corrected execution, forming a supervision signal that captures the human preference. As illustrated in Fig.~\ref{fig:intervention}, these intervention pairs provide targeted corrections in challenging situations and serve as training data for preference-based fine-tuning.

\subsection{Expert intervention data collection in simulated sidewalk environments}
\label{sec:app-human-sim}

We collect expert intervention data in simulation using an ORCA~\cite{van2011reciprocal}-based expert policy. For each episode, we generate a reference route $\tau_{\mathrm{ref}}$ from the start location to the goal and sample intermediate waypoints along the route. The learned policy is deployed in closed loop, and its predicted trajectory is continuously compared with the local reference trajectory. When the average displacement error exceeds a threshold, \ie, $\mathrm{ADE}(\pi_{\theta}, \tau_{\mathrm{ref}}) > 1.0$~m, or when the robot approaches unsafe states such as collision or off-sidewalk regions, the ORCA expert takes over control. During takeover, we record the model trajectory as the negative action and the expert-corrected trajectory as the preferred action. The resulting paired trajectories constitute the intervention dataset $\mathcal{D}_{\mathrm{Int}}$ for finetuning.

\subsection{Details of model architecture}
\label{sec:app-model}

In this section, we provide details of the model architecture used in \ModelName. We adopt a FastViT-based visual encoder~\citep{vasu2023fastvit}. Specifically, we encode the past 10 frames sampled at 20~Hz into frame-level tokens (\ie, one token per frame) via global average pooling, and encode the current frame into patch-level tokens by flattening the backbone feature map. Each token is projected to a $d_{\text{model}}$-dimensional embedding using a lightweight MLP, and the past-frame tokens are further augmented with a sinusoidal time embedding. We additionally use a learnable position embedding over the visual tokens, followed by a 4-layer Transformer with 8 attention heads to extract scene context.

To incorporate goal information, the goal point is embedded by an MLP into a $d_{\text{model}}$-dimensional token and appended to the visual tokens as a conditioning token. To mitigate shortcut learning and improve robustness, we apply stochastic masking over the goal point during training, randomly dropping the goal-point token with a very high probability 0.9.

For action generation, we represent candidate plans using a fixed set of anchor trajectories obtained by K-Means clustering over the training dataset. During training, we apply the anchored flow-matching process to these anchors by sampling a noise level $\tau$ from a Beta distribution and linearly mixing anchor trajectories with Gaussian noise. The noised anchors are encoded by a lightweight action encoder and then refined by an anchored flow-matching Transformer consisting of 4 layers and 8 attention heads, which predicts the flow (velocity field) conditioned on the visual tokens and the goal token via gated cross-attention layers. The refined anchor features are decoded by the MLP heads to predict i) per-step normalized velocities and headings for each anchor mode and ii) a mode score for anchor selection. As shown in Tab.~\ref{tab:arch-hyperparams}, we provide detailed parameters of the model.

\begin{table}[htbp!]
\centering
\small

\begin{tabular}{l|l}
\toprule
\textbf{Parameter} & \textbf{Value} \\
\midrule
RGB Resolution & $352 \times 128$ \\
Observation Sequence  & 11 frames @ 20Hz \\
Backbone & FastViT-MA36~\cite{vasu2023fastvit} \\
Backbone Input Channels & 6 (RGB$_t$ + RGB$_{t-1}$) \\
Backbone Feature Dimension & 1216 \\
Token Dimension ($d_{\text{model}}$) & 1280 \\
Gated Self-Attn Layers & 4 \\
Gated Self-Attn Heads & 8 \\
Dropout & 0.1 \\
Goal Dropout & 0.9 \\
\midrule
Anchors & Fixed (From K-Means) \\
Anchor Modes $K$ & 64 \\
Anchor Horizon & 20 steps @ 5Hz \\
Anchor Dimension & 3D $(\Delta x,\Delta y,\Delta\theta)$ \\
Noise Schedule & $\tau \sim \text{Beta}(1.5, 1.0)$ \\
Time Buckets & 1000 \\
Inference Steps & 4 \\
Gated Cross-Attn Layers & 4 \\
Gated Cross-Attn Heads & 8 \\
Dropout & 0.1 \\
\bottomrule
\end{tabular}
\vspace{1em}

\caption{\textbf{Details of \ModelName\ model.}}
\label{tab:arch-hyperparams}
\end{table}

\subsection{Details of pretraining}
\label{sec:app-pretrain}

As illustrated in Tab.~\ref{tab:pretrain-hyperparams}, we give the details of parameters used in the pretraining stage.

\begin{table}[htbp!]
\centering
\small
\begin{tabular}{l|l}
\toprule
\textbf{Parameter} & \textbf{Value} \\
\midrule
\# Epochs $n_{ep}$ & 50 \\
Batch Size & 32 ($\times$8) \\
weight Decay & $1 \times 10^{-6}$ \\
Learning Rate & $1 \times 10^{-4}$ \\
Optimizer & AdamW \\
LR Schedule & Cosine \\
Scheduler Period & 100 \\
Compute Resources & $8 \times$ NVIDIA PRO 6000 \\
\bottomrule
\end{tabular}
\vspace{1em}
\caption{\textbf{\ModelName\ pretraining.}}
\label{tab:pretrain-hyperparams}
\end{table}

\subsection{Details of finetuning}
\label{sec:app-finetune}

During finetuning, we apply LoRA~\cite{hu2022lora} to the attention projection layers (query/key/value projections). We give the details of parameters used in the finetuning stage as illustrated in Tab.~\ref{tab:finetune-hyperparams}.

\begin{table}[htbp!]
\centering
\small
\begin{tabular}{l|l}
\toprule
\textbf{Parameter} & \textbf{Value} \\
\midrule
\multicolumn{2}{c}{\textbf{Optimization}} \\
\midrule
\# Epochs $n_{ep}$ & 30 \\
Batch Size & 256 \\
Learning Rate & $1 \times 10^{-5}$ \\
Optimizer & AdamW \\
Weight Decay & 0 \\
LR Schedule & Cosine \\
Scheduler Period & 80 \\
Compute Resources & $8 \times$ NVIDIA PRO 6000 \\
\midrule
\multicolumn{2}{c}{\textbf{LoRA}} \\
\midrule
LoRA Rank $r$ & 16 \\
LoRA Scaling $\alpha$ & 16 \\
LoRA Dropout & 0 \\
LoRA Target Modules & Attention projection layers \\
Trainable Parameters & LoRA only (others frozen) \\
\bottomrule
\end{tabular}
\vspace{1em}
\caption{\textbf{\ModelName\ finetuning hyperparameters.}}
\label{tab:finetune-hyperparams}
\end{table}

\subsection{Details of controller}
\label{sec:app-controller}

In practice, we find that this simple PD controller is sufficient for stable deployment since the policy already outputs smooth, goal-directed short-horizon waypoints, and the controller mainly serves as a robust mapping from waypoint space to $(v,w)$ under actuation limits. We convert the predicted short-horizon trajectory into low-level control commands using a lightweight trajectory-based PD controller. Given a predicted waypoint $(\Delta x,\Delta y)$ in the robot local frame over a fixed control interval $\Delta t=1s$, we compute the linear and angular velocities as
$v = \frac{\Delta x}{\Delta t}$ and $w = \frac{\arctan(\Delta y/\Delta x)}{\Delta t}$, with additional safeguards for degenerate cases (\eg, $\Delta x \approx 0$). We further apply standard clipping on velocity ($v_{\text{max}}=1.5\ m/s$) and angular rate ($w_{\text{max}}=0.35\ rad/s$) to satisfy platform constraints, and attenuate the commanded speed based on an estimated local curvature to improve safety in sharp turns ($v=\frac{\Delta x}{\Delta t} \cdot \exp(-0.05\cdot |w|)$ when $|w|\geq0.2$).

\section{Ethics Statement}
\label{sec:ethics}

For real-world deployment, the robots operated under strict safety constraints: maximum velocity and acceleration were limited by both hardware and software. Researchers monitored all experiments and were able to immediately intervene, stop the robot, or take over control to ensure safety. These interventions are explicitly marked as \textbf{Manual} in the figures and video. This protocol ensured that the robot operated under strict safety constraints throughout the experiments.

\end{document}